\documentclass[10pt,twocolumn]{article}

\usepackage[utf8]{inputenc}
\usepackage[T1]{fontenc}
\usepackage{newtxtext,newtxmath} 

\usepackage[a4paper,margin=1in]{geometry}
\usepackage{microtype}           
\usepackage{balance}             

\usepackage{graphicx}
\usepackage{booktabs}
\usepackage{caption}
\usepackage{subcaption}
\usepackage{float}
\usepackage{placeins}            
\usepackage{multirow}
\usepackage{longtable}
\usepackage{natbib}
\setcitestyle{authoryear,round}

\usepackage{multicol}   
\usepackage{booktabs}   
\usepackage{capt-of}

\usepackage{tabularx,booktabs}
\usepackage{ragged2e}                 
\newcolumntype{Y}{>{\RaggedRight\arraybackslash}X}

\usepackage[hidelinks]{hyperref}
\newcommand{\footurl}[2]{\href{#2}{#1}\footnote{\url{#2}}}

\usepackage{enumitem}
\setlist{nosep,leftmargin=*,itemsep=1pt}

\title{\textbf{Simplify-This: A Comparative Analysis of Prompt-Based and Fine-Tuned LLMs}}
\author{
  \textbf{Eilam Cohen \quad Itamar Bul \quad Danielle Inbar \quad Omri Loewenbach} \\
  Tel Aviv University \\
  \small \texttt{eilamc14@gmail.com} \quad \texttt{itamarbul@gmail.com} \quad \texttt{inbar.danielle@gmail.com} \quad \texttt{omrile1@gmail.com}
}
\date{}

\begin{document}
\maketitle

\begin{abstract}
Large language models (LLMs) enable strong text
generation, and in general there is a practical trade-off between \textbf{fine-tuning} and \textbf{prompt engineering}. We introduce \textit{Simplify-This}, a comparative study evaluating both paradigms for \textbf{text simplification} with encoder–decoder LLMs across multiple benchmarks, using a range of evaluation metrics. Fine-tuned models consistently deliver stronger \textbf{structural simplification}, whereas prompting often attains higher \textbf{semantic similarity} scores yet tends to \textbf{copy inputs}. A human evaluation favors fine-tuned outputs overall. We release code, a \textbf{cleaned derivative dataset} used in our study, checkpoints of fine-tuned models, and prompt templates to facilitate reproducibility and future work.
\end{abstract}

\section{Introduction}

Large language models (LLMs) have rapidly transformed natural language processing (NLP). 
Two major paradigms dominate contemporary use: \textit{fine-tuning}, where models are adapted to specific tasks via supervised training on annotated corpora, and \textit{prompt engineering}, where task performance is elicited through carefully designed natural-language instructions without modifying model parameters. 
Fine-tuning has traditionally yielded strong in-domain results, but it requires substantial computational resources and curated datasets. 
Prompting, by contrast, offers flexibility and low cost, yet its effectiveness varies widely across tasks and models.

Among NLP applications, \textbf{text simplification} occupies a distinctive role. 
The task aims to reduce the linguistic complexity of a sentence while preserving its meaning \citep{Shardlow2014}. 
Unlike summarization, which compresses content, simplification targets accessibility by rewriting text to be more comprehensible for non-native speakers, children, or individuals with reading difficulties. 
Despite its apparent simplicity, achieving high-quality simplification is challenging: models must balance readability, fluency, and semantic faithfulness. 
Classical approaches often struggle to avoid trivial copying or excessive distortion, while automatic metrics capture complementary yet incomplete aspects of quality (e.g., structural, readability, and semantic similarity measures).

Foundational advances in pre-trained sequence-to-sequence models such as T5 \citep{2020t5} and BART \citep{lewis-etal-2020-bart} have significantly improved the state of the art in text generation, and subsequent work has applied these architectures to simplification with encouraging results \citep{martin2021mussmultilingualunsupervisedsentence, basu2023med, kew-etal-2023-bless}. 
Instruction-tuned variants such as Flan-T5 \citep{chung2022scalinginstructionfinetunedlanguagemodels} further raise the question of whether prompting alone can rival fine-tuning in this task. 
However, systematic comparisons between the two paradigms in text simplification remain limited.

In this paper, we present \textbf{Simplify-This}, a comparative study of fine-tuning versus prompt-based approaches for text simplification. 
We evaluated a diverse set of pre-trained models across three commonly used datasets: ASSET \citep{alva-manchego-etal-2020-asset}, Med-EASi \citep{basu2023med}, and OneStopEnglish \citep{vajjala-lucic-2018-onestopenglish}. 
Our analysis integrates automatic metrics with human preference judgments, providing a comprehensive view of the trade-offs between edit-sensitive and faithfulness-oriented criteria. 
Beyond reporting results, we highlight when prompting can approximate fine-tuning, where fine-tuning remains indispensable, and how dataset domain may influence the relative success of each paradigm.

Overall, our contributions are threefold:
\begin{itemize}
    \item We conduct, to the best of our knowledge, among the first systematic head-to-head comparison of fine-tuning and prompting for text simplification across multiple datasets, models, and evaluation methods.
    \item We release cleaned data (\textit{WikiLarge-Clean}), code, prompts, and fine-tuned checkpoints to support reproducibility and further research.
    \item We provide both quantitative and qualitative analyses, including human evaluation, to contextualize the strengths and limitations of each approach.
\end{itemize}
All resources are publicly available (see Section ~\ref{sec:release}).

\section{Related Work}

\subsection{Text simplification with LLMs}  
The emergence of instruction-tuned LLMs has reshaped English text simplification (TS). The BLESS benchmark systematically evaluated 44 off-the-shelf LLMs under few-shot prompting across Wikipedia, news, and medical domains, showing that the best prompted LLMs can perform comparably to state-of-the-art supervised TS baselines without task-specific fine-tuning \citep{kew-etal-2023-bless}. Building on this, \citet{WuArase2025} conducted an error-based human evaluation and found that GPT-4 produces fewer erroneous simplifications than prior fine-tuned models, though challenges remain in lexical paraphrasing and content fidelity. They further caution that commonly used automatic metrics can be insensitive among high-quality outputs.

\subsection{Prompting vs. fine-tuning}  
Across related rewriting tasks, comparisons between prompting and fine-tuning reveal a pragmatic trade-off. In the BioLaySumm shared task, fine-tuning a Longformer-based model outperformed GPT-4 prompting when the full training set was available, whereas GPT-4 prompting excelled in low-data settings \citep{to-etal-2024-deakinnlp}. This suggests prompting is often effective in scarce-data regimes, while fine-tuning often excels when large, domain-specific corpora exist. For document-level TS, \citet{Fang2025ProgDS} propose a progressive multi-stage prompting method (ProgDS), which decomposes simplification into discourse-, topic-, and lexical-level phases. ProgDS significantly outperforms direct one-shot prompting and smaller fine-tuned models, underscoring the potential of structured prompt design when simplifying long texts.

\subsection{Datasets and domains}  
Prior work on simplification commonly evaluates models on ASSET \citep{alva-manchego-etal-2020-asset}, 
Newsela \citep{Xu2015Newsela}, 
and Med-EASi \citep{basu2023med}. 
The BLESS benchmark explicitly included these domains and reported that strong LLMs generalize across them, 
though with domain-specific error patterns \citep{kew-etal-2023-bless}. 
In addition, many studies rely on the WikiLarge corpus \citep{zhang-lapata-2017-sentence, Xu2016OptimizingSM},
though several limitations have been documented \citep{alva-manchego-etal-2020-asset}.

\subsection{Metrics and control}  
The field routinely reports SARI \citep{Xu2016OptimizingSM}, 
FKGL \citep{kincaid1975derivation}, BERTScore, \citep{DBLP:journals/corr/abs-1904-09675} and LENS \citep{maddela2023lenslearnableevaluationmetric}. 
\citet{WuArase2025} emphasize that automatic metrics are sometimes insensitive among high-quality outputs, 
motivating human evaluation as a complementary measure. 
Beyond generic metrics, readability-controlled rewriting remains challenging: 
\citet{farajidizaji-etal-2024-possible} show that LLMs can be prompted to adjust complexity up or down 
but often fail to reach precise grade levels, with readability still correlating strongly with the original input.

\section{Methodology}
\subsection{Task}
We study sentence-level text simplification: given an input sentence x, generate an output y that preserves meaning and is easier to understand while improving readability.

\subsection{Models}
For this study, we focused on \textbf{sequence-to-sequence (seq2seq) architectures}, as they are well-suited to natural language generation tasks where the goal is to transform an input sequence into a modified but semantically related output. Seq2seq models have proven highly effective in a wide range of NLP applications, including machine translation \citep{bahdanau2016neuralmachinetranslationjointly}, summarization \citep{see-etal-2017-get}, and dialogue systems \citep{vinyals2015neuralconversationalmodel}. Their encoder–decoder structure allows the model to capture the semantic content of an input sentence and generate a controlled output, making them especially appropriate for \textbf{text simplification}, where meaning must be preserved while surface form is simplified \citep{zhang-lapata-2017-sentence, Xu2016OptimizingSM}.

In practice, one notable example is \textbf{MUSS}, a BART-based model fine-tuned specifically for text simplification \citep{martin2021mussmultilingualunsupervisedsentence}. MUSS demonstrated that pretrained seq2seq models, when adapted through targeted fine-tuning, can outperform previous supervised approaches and achieve state-of-the-art results on benchmarks such as WikiLarge

Our project was also constrained by computational and financial resources. Training large-scale models is prohibitive in terms of cost and compute time. To ensure we could conduct full fine-tuning experiments within our limited budget, we deliberately selected open-source models under 1 billion parameters from \href{https://huggingface.co}{Hugging Face}. This allowed us to run multiple experiments efficiently on limited hardware while still comparing competitive and diverse architectures.

We therefore selected a representative set of \textbf{open-source encoder--decoder models} from the Hugging Face Hub \citep{wolf2020huggingfacestransformersstateoftheartnatural}. Our study covers the following:

\begin{itemize}
    \item \textbf{BART-base (139M)} and \textbf{BART-large (406M)} - denoising autoencoder seq2seq models shown effective for text generation and summarization \citep{lewis-etal-2020-bart}.
    \item \textbf{T5-base (223M)} and \textbf{T5-large (738M)} - the ``Text-to-Text Transfer Transformer'' framework, which frames all NLP tasks as sequence-to-sequence \citep{2020t5}.
    \item \textbf{Flan-T5-base (248M)} and \textbf{Flan-T5-large (783M)} - instruction-tuned versions of T5 designed to improve generalization to unseen tasks \citep{chung2022scalinginstructionfinetunedlanguagemodels}.
    \item \textbf{Pegasus-large (571M)} and \textbf{Pegasus-xsum (570M)} - models pretrained with a gap-sentence objective tailored for abstractive summarization and transfer \citep{zhang2020pegasuspretrainingextractedgapsentences}. The \textit{XSum} variant is further trained on the Extreme Summarization dataset \citep{narayan-etal-2018-dont}, which consists of highly abstractive one-sentence summaries of BBC news articles.
    \item \textbf{ProphetNet-large-uncased-cnndm (391M)} - introducing a novel self-supervised objective of future $n$-gram prediction for better sequence modeling \citep{qi2020prophetnetpredictingfuturengram}. 
This variant is further pretrained and evaluated on the CNN/DailyMail summarization corpus \citep{hermann2015teachingmachinesreadcomprehend,see-etal-2017-get}, a benchmark designed to assess abstractive summarization. 
\end{itemize}

This selection provides a diverse range of architectures and training paradigms, enabling us to assess how fine-tuning and prompt engineering perform across different seq2seq foundations.

\subsection{Datasets}

\subsubsection{Evaluation Datasets}

The selection of evaluation datasets in this study was guided by the BLESS benchmark \citep{kew-etal-2023-bless}, which provides a comprehensive framework for evaluating text simplification systems. Following these guidelines, we included datasets that offer diverse perspectives on simplification quality and allow for multi-faceted evaluation. 

In addition to the core benchmarks recommended in BLESS, we also incorporated \textbf{OneStopEnglish}, motivated by the work of \cite{vajjala-lucic-2018-onestopenglish}, who introduced this corpus of news articles rewritten at three proficiency levels 
and demonstrated its value for assessing simplification in educational (EFL) settings. The final set of evaluation datasets thus consists of \textbf{ASSET} \citep{alva-manchego-etal-2020-asset}, \textbf{Med-EASi} \citep{basu2023med}, and \textbf{OneStopEnglish} \citep{vajjala-lucic-2018-onestopenglish}.

\paragraph{ASSET} 
ASSET \citep{alva-manchego-etal-2020-asset} is a benchmark dataset created to overcome key limitations of earlier corpora such as WikiLarge. In contrast to single-reference datasets, ASSET provides ten human-written simplifications for each of its 2,359 original sentences. This design enables more robust evaluation by capturing the wide range of valid simplification strategies, including both lexical and syntactic transformations. Furthermore, the multiple references reduce the bias toward any single rewriting style, allowing metrics such as SARI to better reflect performance. In our experiments, we rely exclusively on the \textbf{test subset} (359 instances) of ASSET, which has become a standard benchmark for evaluating the quality of text simplification systems.

\paragraph{Med-EASi} 
Med-EASi \citep{basu2023med} is a more recent finely annotated dataset for medical text simplification. The corpus contains approximately 1,979 expert–simple text pairs and is annotated with four types of edit operations — \textit{elaboration}, \textit{replacement}, \textit{deletion}, and \textit{insertion} — to support controllable simplification. The source texts are drawn from medical resources (e.g., Merck Manuals) and SimpWiki, and the dataset was created with medical expert oversight to ensure fidelity in domain-specific rewriting. For evaluation we use the released \textbf{test subset} (300 instances).

\paragraph{OneStopEnglish}
OneStopEnglish \citep{vajjala-lucic-2018-onestopenglish} contains news articles rewritten by professional teachers at three proficiency levels: \textit{advanced}, \textit{intermediate}, and \textit{elementary}. In our evaluation, we operate at the sentence level, using the \textbf{advanced} sentences as sources and their aligned \textbf{elementary} counterparts as targets. The resulting alignment comprises 2,178 sentence pairs, enabling a fine-grained assessment of simplification quality from complex to beginner-friendly language.

\subsubsection{Training Dataset}
\label{sec:training-dataset}

For fine-tuning, we rely on \textbf{WikiLarge-Clean}, a dataset we constructed specifically for this study, building on the widely used \textbf{WikiLarge} corpus \citep{zhang-lapata-2017-sentence,Xu2016OptimizingSM}. WikiLarge has served as the de facto training resource for text simplification, combining sentence pairs from Simple English Wikipedia, the PWKP dataset, and the TurkCorpus. Despite its broad adoption, several limitations have been documented: the presence of noisy alignments, numerous near-duplicate pairs, sentences with trivial identity mappings, and inconsistent preprocessing \citep{alva2020survey}. These issues reduce the reliability of fine-tuning and can bias evaluation metrics. Additionally, due to limited computational and financial resources, we opted to reduce the size of the original WikiLarge training pool—originally containing 296,402 automatically-aligned sentence pairs \citep{zhang-lapata-2017-sentence}.

\textbf{WikiLarge-Clean} was created through a multi-step preprocessing pipeline aimed at producing a higher-quality corpus. All filtering thresholds were applied over whitespace-delimited tokens (i.e., without model-specific tokenization). The cleaning procedure followed the order defined in our preprocessing code (see Appendix~\ref{sec:appendix-dataset} for full details):

\begin{itemize}
    \item \textbf{Length constraints}
    \item \textbf{Compression ratio}
    \item \textbf{Near-identity (lexical Jaccard similarity) filtering}
    \item \textbf{Deduplication}
\end{itemize}

\begin{table}[t]
\centering
\caption{Before and after statistics and scores for the train split of WikiLarge/Clean. *(no trailing period)}
\label{tab:cleaning-train2}
\begin{tabular}{lcc}
\toprule
Stat / Metric & Before & After \\
\midrule
Size & 296{,}402 & 123{,}862 \\
Mean Compression Ratio & 0.88 & 0.70 \\
Mean Source Length & 25.17 & 26.49 \\
Mean Target Length & 18.51 & 18.29 \\
FKGL $\downarrow$ & 9.24 & 8.83 \\
BERTScore $\uparrow$ & 46.27 & 52.18 \\
LENS-SALSA $\uparrow$ & 49.13 & 54.44 \\
Near-identical* & 1214 & 0 \\
\bottomrule
\end{tabular}
\end{table}

The resulting dataset contains approximately 124k high-quality sentence pairs, split into training, validation, and test subsets (derived from the original train/valid/test split). This makes \textit{WikiLarge-Clean} suitable for both large-scale fine-tuning and systematic evaluation.

As shown in Table~\ref{tab:cleaning-train2}, these filtering steps lead to a more balanced dataset: 
the mean compression ratio decreases from 0.88 to 0.70, indicating reduced copying behavior, while average source and target lengths remain stable. 
Readability, as measured by FKGL, improves slightly \citep{flesch1948yardstick,kincaid1975derivation}, and quality-oriented metrics such as BERTScore \citep{DBLP:journals/corr/abs-1904-09675} and LENS-SALSA (a variant of LENS designed for reference-free evaluation; \citet{maddela2023lenslearnableevaluationmetric,heineman2023dancingsuccessfailureeditlevel}) show consistent gains. 
Notably, the number of near-identical pairs drops to zero, confirming that the cleaned dataset eliminates trivial simplifications.

\subsection{Fine-Tuning Setup}
\label{sec:ft-setup}
\paragraph{Objective and Training Regime}
We perform \textbf{full fine-tuning} (all parameters) with a standard seq2seq cross-entropy objective, using Hugging Face \texttt{Seq2SeqTrainer}. Early stopping monitors validation performance and keeps the best checkpoint. We adopt dynamic padding during training/evaluation.

\paragraph{Preprocessing}
Each source sentence is prefixed with \texttt{``Simplify: ''} and both inputs and labels are tokenized with the base model’s tokenizer. This input convention is kept consistent across all models.

\paragraph{Hardware and Memory Controls}
Experiments run on a single NVIDIA L4 GPU (Google Colab). To fit large encoder–decoder models, we enable \textit{gradient checkpointing} and disable the KV cache during training.

\paragraph{Hyperparameters.}
For each model we conducted several fine-tuning runs with small variations in hyperparameters (e.g., learning rate, batch size etc.) in order to identify the most effective configuration. While we do not report the exact final settings of each individual run, the training followed a consistent regime and was based on standard practices for encoder–decoder LLMs. We report in Appendix~\ref{sec:appendix-finetune} the typical default configuration, which represents the regime under which most models were fine-tuned.

\paragraph{Training Data}
All models are fine-tuned on our \textbf{WikiLarge-Clean} corpus (cleaned WikiLarge-style pairs) described in \S\ref{sec:training-dataset}

\paragraph{Environmental Note}
All fine-tuning experiments were conducted on Google Cloud’s managed GPU service, typically provisioning a single NVIDIA L4 accelerator. The exact resource allocation and utilization details are abstracted by the provider, so we cannot precisely quantify runtime efficiency or carbon footprint. Nonetheless, the practical runtime per model was on the order of 5–10 GPU hours, suggesting relatively modest computational demands compared to large-scale pretraining.

\subsection{Prompt Engineering Setup}
We implemented ten prompt templates (P1--P10) covering major strategies such as zero-shot instructions, multi-shot demonstrations, chain-of-thought reasoning, lexical simplification, sentence splitting, readability control, and content preservation. We added P0 as a control prompt, which is identical to the fine-tune preprocessing (``Simplify: <src>``).
Complete templates and references are provided in Appendix~\ref{sec:appendix-prompts}).

\subsection{Evaluation Metrics}

Our choice of metrics and evaluation protocol is informed by prior simplification benchmarks and meta-studies, most notably the BLESS benchmark \citep{kew-etal-2023-bless}. Each metric captures a different aspect of simplification quality, ranging from content preservation to grammaticality, fluency, and complexity reduction.
Specifically, we report SARI \citep{Xu2016OptimizingSM}, FKGL \citep{flesch1948yardstick,kincaid1975derivation}, BERTScore \citep{DBLP:journals/corr/abs-1904-09675}, and LENS \citep{maddela2023lenslearnableevaluationmetric}. In addition, we compute an Identical Ratio (id\_ratio), the fraction of outputs identical to the input, as a sanity check against trivial copying. Formal definitions and implementation details are provided in Appendix~\ref{sec:appendix-metrics}.

\subsection{Human Evaluation}
In addition to automatic metrics, we conducted a blinded human preference study. 
For each \textit{model $\times$ dataset} configuration, we paired, per source, the best prompt-based output (by SARI, tie-broken by LENS) against the fine-tuned output; 
pairs with copying/prompt leakage were filtered. if multiple ``good'' candidates remained, one was selected at random.

The resulting pairs were presented in Qualtrics
(see Appendix~\ref{sec:qualtrics} for the survey link). 
Each trial displayed the source sentence along with two candidate simplifications (order randomized, provenance hidden). 
Raters were asked to select the better simplification, with a ``same'' option available but discouraged if a preference existed. 
Responses were scored as +1 (Prompt), $-1$ (Fine-tuned), or 0 (Same). 
Thirteen raters each evaluated 20 pairs, yielding 260 judgments in total.

\section{Results}
\subsection{Quantitative Results}
\subsubsection{Metric Evaluation Results}
For each dataset, we report three configurations per model: \textbf{FT} (fine-tuned), \textbf{P-SARI} (prompt-best by SARI), and \textbf{P-LENS} (prompt-best by LENS). 
This compact view highlights the trade-offs between edit-sensitive and conservatism-tolerant metrics. 
Prompts flagged as copy-heavy (high identical ratio) are marked with $\dagger$. 
Full per-prompt results for all models and prompts are provided in Appendix~\ref{app:full_tables}.

\begin{table*}[t]
\centering
\caption{ASSET: per-model comparison of \textit{FT}, \textit{P-SARI}, and \textit{P-LENS}.
Identical ratio is case-sensitive for all systems and case-insensitive for ProphetNet (ci).
Copy-heavy systems (Identical ratio $>0.50$) are marked with $\dagger$.}
\label{tab:asset-threeway}
\small
\begin{tabular}{l l c c c c c c}
\toprule
Model & Variant & Prompt\# & Identical ratio & SARI $\uparrow$ & FKGL $\downarrow$ & BERTScore $\uparrow$ & LENS $\uparrow$ \\
\midrule
\multirow{3}{*}{BART-base}
  & FT         & --  & 0.00              & \textbf{36.13} & \textbf{8.50} & 85.57 & 43.88 \\
  & P-SARI   & 9   & 0.70$^\dagger$    & 26.88 & 9.38 & 86.59 & 51.58 \\
  & P-LENS   & 0   & 0.93$^\dagger$    & 21.44 & 10.02 & \textbf{90.80} & \textbf{60.05} \\
\midrule
\multirow{3}{*}{BART-large}
  & FT         & --  & 0.01              & \textbf{37.96} & \textbf{7.87} & 84.64 & 44.51 \\
  & P-SARI   & 9   & 0.69$^\dagger$    & 27.14 & 9.29 & 86.31 & 51.48 \\
  & P-LENS   & 0   & 0.92$^\dagger$    & 21.81 & 9.78 & \textbf{90.38} & \textbf{59.55} \\
\midrule
\multirow{3}{*}{T5-base}
  & FT         & --  & 0.20              & \textbf{35.38} & 8.45 & \textbf{86.85} & \textbf{57.43} \\
  & P-SARI   & 9   & 0.01              & 35.05 & \textbf{6.96} & 22.16 & 30.00 \\
  & P-LENS   & 0   & 0.27              & 29.11 & 8.61 & 81.96 & 49.27 \\
\midrule
\multirow{3}{*}{T5-large}
  & FT         & --  & 0.22              & 35.41 & 8.54 & \textbf{87.04} & \textbf{59.81} \\
  & P-SARI   & 2   & 0.00              & \textbf{36.18} & 4.79 & 44.43 & 5.07 \\
  & P-LENS   & 3   & 0.00              & 32.71 & \textbf{1.41} & 45.37 & 10.99 \\
\midrule
\multirow{3}{*}{Flan-T5-base}
  & FT         & --  & 0.01              & \textbf{37.92} & 8.22 & 84.40 & 48.29 \\
  & P-SARI   & 0   & 0.13              & 36.23 & \textbf{6.91} & 58.13 & 41.14 \\
  & P-LENS   & 4   & 0.31              & 33.37 & 8.62 & \textbf{88.67} & \textbf{64.53} \\
\midrule
\multirow{3}{*}{Flan-T5-large}
  & FT         & --  & 0.02              & \textbf{37.91} & \textbf{7.86} & 84.03 & 47.87 \\
  & P-SARI   & 4   & 0.22              & 36.31 & 8.28 & \textbf{86.79} & \textbf{66.31} \\
  & P-LENS   & 4   & 0.22              & 36.31 & 8.28 & \textbf{86.79} & \textbf{66.31} \\
\midrule
\multirow{3}{*}{Pegasus-large}
  & FT         & --  & 0.24              & \textbf{35.67} & \textbf{8.79} & 86.25 & \textbf{61.52} \\
  & P-SARI   & 2   & 0.00              & 26.37 & 9.64 & 19.66 & 30.97 \\
  & P-LENS   & 0   & 0.86$^\dagger$    & 23.33 & 10.24 & \textbf{88.74} & 58.48 \\
\midrule
\multirow{3}{*}{Pegasus-xsum}
  & FT         & --  & 0.29              & 33.80 & 9.23 & \textbf{87.54} & 62.46 \\
  & P-SARI   & 8   & 0.06              & \textbf{34.52} & 8.08 & 37.17 & 57.97 \\
  & P-LENS   & 3   & 0.01              & 28.63 & \textbf{5.36} & 12.34 & \textbf{65.49} \\
\midrule
\multirow{3}{*}{\shortstack[l]{ProphetNet-large-\\uncased-cnndm}}
  & FT         & --  & 0.11 (ci)         & \textbf{38.01} & 7.70 & \textbf{67.82} & \textbf{60.85} \\
  & P-SARI   & 0   & 0.16 (ci)         & 37.76 & \textbf{5.75} & 62.30 & 51.60 \\
  & P-LENS   & 1   & 0.22 (ci)         & 34.58 & 8.00 & 64.85 & 51.63 \\
\bottomrule
\end{tabular}
\end{table*}

\begin{table*}[t]
\centering
\caption{Med-EASi: per-model comparison of \textit{FT}, \textit{P-SARI}, and \textit{P-LENS}.
Identical ratio is case-sensitive for all systems and case-insensitive for ProphetNet (ci).
Copy-heavy systems (Identical ratio $>0.50$) are marked with $\dagger$.}
\label{tab:medeasi-threeway}
\small
\begin{tabular}{llcccccc}
\toprule
Model & Variant & Prompt\# & Identical ratio & SARI $\uparrow$ & FKGL $\downarrow$ & BERTScore $\uparrow$ & LENS $\uparrow$ \\
\midrule
\multirow{3}{*}{BART-base}
  & FT         & --  & 0.02            & \textbf{33.47} & 10.49 & 44.16 & 35.33 \\
  & P-SARI   & 9   & 0.59$^\dagger$  & 33.44 & \textbf{10.14} & 44.96 & 41.14 \\
  & P-LENS   & 0   & 0.86$^\dagger$  & 24.68 & 11.20 & \textbf{47.63} & \textbf{49.44} \\
\midrule
\multirow{3}{*}{BART-large}
  & FT         & --  & 0.02            & \textbf{36.25} & \textbf{10.21} & 44.49 & 36.08 \\
  & P-SARI   & 9   & 0.56$^\dagger$  & 33.77 & 10.08 & 45.23 & 41.96 \\
  & P-LENS   & 0   & 0.84$^\dagger$  & 24.76 & 11.10 & \textbf{47.96} & \textbf{49.46} \\
\midrule
\multirow{3}{*}{T5-base}
  & FT      & -- & 0.22 & 33.43 & \textbf{10.59} & \textbf{44.97} & \textbf{45.69} \\
  & P-SARI  & 9  & 0.01 & \textbf{38.00} & 7.76 & 6.48 & 32.96 \\
  & P-LENS  & 0  & 0.23 & 32.45 & 10.30 & 42.04 & 40.17 \\
\midrule
\multirow{3}{*}{T5-large}
  & FT      & -- & 0.20 & 33.22 & \textbf{10.53} & \textbf{44.87} & \textbf{48.53} \\
  & P-SARI  & 7  & 0.00 & \textbf{38.22} & 3.86 & 10.99 & 3.54 \\
  & P-LENS  & 3  & 0.00 & 34.55 & 2.22 & 20.05 & 7.55 \\
\midrule
\multirow{3}{*}{Flan-T5-base}
  & FT         & --  & 0.04            & \textbf{36.24} & 9.68 & 42.63 & 38.44 \\
  & P-SARI   & 0   & 0.19            & 35.29 & \textbf{7.95} & 27.58 & 35.86 \\
  & P-LENS   & 6   & 0.27            & 33.88 & 9.73 & \textbf{44.17} & \textbf{60.03} \\
\midrule
\multirow{3}{*}{Flan-T5-large}
  & FT         & --  & 0.06            & \textbf{36.62} & \textbf{9.38} & 43.29 & 38.79 \\
  & P-SARI   & 4   & 0.26            & 35.62 & 9.56 & \textbf{44.70} & \textbf{61.84} \\
  & P-LENS   & 4   & 0.26            & 35.62 & 9.56 & \textbf{44.70} & \textbf{61.84} \\
\midrule
\multirow{3}{*}{Pegasus-large}
  & FT         & --  & 0.45            & \textbf{28.56} & \textbf{11.09} & 49.88 & 54.39 \\
  & P-SARI   & 8   & 0.00            & 24.92 & 16.69 & 33.08 & 12.35 \\
  & P-LENS   & 10  & 0.87$^\dagger$  & 22.37 & 11.74 & \textbf{51.66} & \textbf{55.26} \\
\midrule
\multirow{3}{*}{Pegasus-xsum}
  & FT         & --  & 0.45            & 27.09 & 11.39 & \textbf{49.94} & 55.35 \\
  & P-SARI   & 8   & 0.10            & \textbf{29.17} & 10.15 & 22.65 & 47.99 \\
  & P-LENS   & 3   & 0.02            & 24.09 & \textbf{6.81} & 8.80  & \textbf{60.22} \\
\midrule
\multirow{3}{*}{\shortstack[l]{ProphetNet-large-\\uncased-cnndm}}
  & FT         & --  & 0.11 (ci)       & \textbf{36.45} & 9.33 & \textbf{40.18} & \textbf{55.99} \\
  & P-SARI   & 0   & 0.16 (ci)       & 36.12 & \textbf{7.34} & 35.94 & 48.68 \\
  & P-LENS   & 1   & 0.22 (ci)       & 33.69 & 9.66 & 38.24 & 48.74 \\
\bottomrule
\end{tabular}
\end{table*}

\begin{table*}[t]
\centering
\caption{OneStopEnglish: per-model comparison of \textit{FT}, \textit{P-SARI}, and \textit{P-LENS}.
Identical ratio is case-sensitive for all systems and case-insensitive for ProphetNet (ci).
Copy-heavy systems (Identical ratio $>0.50$) are marked with $\dagger$.}
\label{tab:ose-threeway}
\small
\begin{tabular}{llcccccc}
\toprule
Model & Variant & Prompt\# & Identical ratio & SARI $\uparrow$ & FKGL $\downarrow$ & BERTScore $\uparrow$ & LENS $\uparrow$ \\
\midrule
\multirow{3}{*}{BART-base}
  & FT        & --  & 0.00 & \textbf{37.45} & \textbf{8.08} & 75.46 & 41.18 \\
  & P-SARI  &  9  & 0.62$^\dagger$ & 32.21 & 8.24 & 76.02 & 49.72 \\
  & P-LENS  &  0  & 0.93$^\dagger$ & 27.36 & 9.26 & \textbf{81.10} & \textbf{59.80} \\
\midrule
\multirow{3}{*}{BART-large}
  & FT        & --  & 0.00 & \textbf{39.99} & \textbf{7.62} & 76.09 & 43.28 \\
  & P-SARI  &  9  & 0.61$^\dagger$ & 32.32 & 8.24 & 75.50 & 49.69 \\
  & P-LENS  &  0  & 0.90$^\dagger$ & 27.60 & 9.24 & \textbf{80.55} & \textbf{60.11} \\
\midrule
\multirow{3}{*}{T5-base}
  & FT        & --  & 0.31 & \textbf{37.51} & 8.14 & \textbf{76.70} & \textbf{56.97} \\
  & P-SARI  &  9  & 0.02 & 30.87 & \textbf{6.94} & 17.85 & 30.72 \\
  & P-LENS  &  0  & 0.32 & 33.63 & 7.58 & 73.81 & 50.35 \\
\midrule
\multirow{3}{*}{T5-large}
  & FT        & --  & 0.32 & \textbf{39.43} & 8.16 & \textbf{78.24} & \textbf{60.31} \\
  & P-SARI  &  7  & 0.00 & 35.92 & 2.51 & 28.62 & 4.27 \\
  & P-LENS  &  3  & 0.00 & 34.43 & \textbf{0.30} & 40.81 & 10.28 \\
\midrule
\multirow{3}{*}{Flan-T5-base}
  & FT        & --  & 0.02 & \textbf{37.73} & 7.57 & 72.84 & 45.97 \\
  & P-SARI  &  4  & 0.21 & 37.24 & 7.79 & 74.68 & 63.78 \\
  & P-LENS  &  4  & 0.21 & 37.24 & 7.79 & 74.68 & \textbf{63.78} \\
\midrule
\multirow{3}{*}{Flan-T5-large}
  & FT        & --  & 0.04 & \textbf{38.38} & 7.64 & 74.23 & 46.45 \\
  & P-SARI  &  4  & 0.20 & 38.10 & 7.55 & 73.40 & 65.02 \\
  & P-LENS  &  4  & 0.20 & 38.10 & \textbf{7.55} & 73.40 & \textbf{65.02} \\
\midrule
\multirow{3}{*}{Pegasus-large}
  & FT        & --  & 0.41 & \textbf{36.89} & 8.42 & 77.79 & \textbf{61.13} \\
  & P-SARI  &  8  & 0.00 & 29.43 & 16.39 & 38.96 & 12.42 \\
  & P-LENS  &  0  & 0.93$^\dagger$ & 26.95 & 9.49 & \textbf{80.89} & 59.32 \\
\midrule
\multirow{3}{*}{Pegasus-xsum}
  & FT        & --  & 0.40 & \textbf{37.07} & 8.66 & \textbf{77.77} & 60.97 \\
  & P-SARI  &  8  & 0.06 & 28.31 & 8.31 & 27.34 & 57.04 \\
  & P-LENS  &  3  & 0.02 & 21.39 & \textbf{5.51} & 9.74 & \textbf{65.54} \\
\midrule
\multirow{3}{*}{\shortstack[l]{ProphetNet-large-\\uncased-cnndm}}
  & FT        & --  & 0.13 (ci) & \textbf{39.17} & \textbf{7.00} & \textbf{65.22} & \textbf{61.53} \\
  & P-SARI  &  1  & 0.14 (ci) & 35.60 & 7.05 & 58.27 & 50.91 \\
  & P-LENS  &  1  & 0.14 (ci) & 35.60 & 7.05 & 58.27 & 50.91 \\
\bottomrule
\end{tabular}
\end{table*}

Across all three datasets (Tables~\ref{tab:asset-threeway}, \ref{tab:medeasi-threeway}, and \ref{tab:ose-threeway}), the general trend is that \textbf{fine-tuned models achieve the highest SARI scores}, while \textbf{prompt-based variants often obtain higher LENS or BERTScore}. 
However, prompt configurations with high LENS are frequently copy-heavy, as indicated by Identical ratio values above 0.50 ($^\dagger$). 
This reinforces the importance of reporting Identical ratio as a sanity check: gains in faithfulness-oriented metrics do not necessarily reflect genuine simplification. 
At the same time, we also observe \textbf{non-trivial cases where high BERTScore or LENS values occur without extensive copying}, particularly in summarization-pretrained models such as Pegasus and ProphetNet, which tend to preserve semantic fidelity via paraphrasing rather than verbatim overlap, and occasionally in Flan-T5 (e.g., on OSE).

\paragraph{ASSET} 
Because ASSET includes multiple human references \citep{alva-manchego-etal-2020-asset}, SARI scores tend to be relatively high and are most sensitive to structural simplification. 
Here, fine-tuned models consistently outperform prompt-based ones in SARI. 
Prompt-based variants occasionally match or exceed in LENS (e.g., BART and Pegasus), but these cases are usually accompanied by high Identical ratio values, reflecting copying rather than true simplification.

\paragraph{Med-EASi} 
Medical texts are inherently complex, leading to overall higher FKGL scores across models \citep{basu2023med}. 
The fine-tuned models again dominate in SARI, while prompt-based ones often maximize LENS—but nearly always with $^\dagger$, confirming copy-heavy outputs. 
This suggests that prompt engineering alone is insufficient for domain-specific corpora with specialized vocabulary, where fine-tuning provides clearer gains.

\paragraph{OneStopEnglish} 
In OSE, which contains shorter and more pedagogical texts \citep{vajjala-lucic-2018-onestopenglish}, prompt-based models are somewhat more competitive. 
For example, Flan-T5 with Prompt \#4 reaches strong performance on both SARI and LENS without necessarily triggering the copy-heavy marker. 
Still, fine-tuning remains the only strategy that yields consistently high SARI across all models, highlighting its importance even when prompts appear viable.

\paragraph{Cross-dataset synthesis per model.}
\textit{BART} (base/large) shows the most balanced trade-off: fine-tuning achieves the best SARI, while prompt-based runs often yield the best LENS, albeit with high copying. 
\textit{T5} (base/large) generally requires fine-tuning; in ASSET, T5-large with Prompt \#2 is a rare exception where a prompt equals or surpasses FT in SARI, but this is not observed elsewhere. 
\textit{Flan-T5} highlights dataset sensitivity: in OSE, prompt engineering can be competitive, while in ASSET and Med-EASi fine-tuning is clearly superior. 
\textit{Pegasus} shows mixed patterns: Pegasus-large fine-tuning dominates both SARI and LENS in some datasets, whereas Pegasus-xsum displays the usual PB=LENS vs.\ FT=SARI trade-off. 
Finally, \textit{ProphetNet} stands out: despite being pretrained on summarization, its fine-tuned version is consistently strong across all metrics, including LENS, and its case-insensitive Identical ratio values remain relatively low—indicating that improvements are not driven solely by copying.

Overall, the results underscore a clear trend: \textbf{fine-tuning yields robust simplification (SARI)}, whereas \textbf{prompting can optimize lexical similarity metrics (LENS, BERTScore)} but is prone to superficial copying. 
While exceptions exist (e.g., Flan-T5 on OSE, T5-large on ASSET), and certain models (notably Pegasus and Flan-T5) demonstrate high similarity metrics without purely relying on copying, these remain limited. 
Thus, fine-tuning continues to be the most reliable strategy for achieving genuine simplification across diverse datasets.

\subsubsection{Human Evaluation Results}
Overall, raters preferred fine-tuned outputs more often (122 choices, 46.9\%) than prompt-engineered outputs (82 choices, 31.5\%), with 56 judgments (21.5\%) marked as ``same''. 
Excluding ties, the prompt win-rate was 40.2\% (95\% CI [33.7, 47.0]). 
An exact binomial test against chance (50\%) confirmed a significant fine-tuned advantage ($p = 0.0062$). 
Including ties as neutral, the mean per-trial score was $-0.154$, corresponding to an overall split of approximately 42\% prompt vs.\ 58\% fine-tuned. 
These results indicate that human raters consistently preferred fine-tuned simplifications over prompt-based ones.

\section{Discussion}
The findings highlight a persistent trade-off between \textit{edit-sensitive} metrics such as SARI and \textit{conservatism-tolerant} metrics such as LENS and BERTScore. 
Fine-tuned models typically achieve higher SARI, confirming their ability to produce genuine structural simplification, while prompt-based models often excel in faithfulness-oriented scores but at the cost of superficial copying. 
This dichotomy reflects a broader methodological question: whether simplification should prioritize measurable edits or semantic preservation, and how to balance the two.

Another dimension concerns efficiency. 
Fine-tuning requires substantial compute resources and domain-specific training data, but yields stable improvements across datasets and models. 
Prompt engineering, by contrast, is far more lightweight and adaptable, making it attractive in low-resource or rapid-deployment settings, yet its effectiveness is highly variable across datasets. 
In practice, this suggests a cost–benefit trade-off, where fine-tuning is preferable when stability and robustness are essential, while prompting may suffice for exploratory or low-stakes applications.

Importantly, the human evaluation corroborates these findings. 
Raters consistently preferred fine-tuned outputs over prompt-based ones, even though the prompt results were \textit{post hoc} filtered to showcase their strongest-performing generations. 
This strengthens the conclusion that fine-tuning produces qualitatively superior simplifications, not only higher metric scores.

Finally, several limitations constrain the interpretation of our results. 
Our evaluation spans three widely used datasets but does not cover lower-resource languages, noisy user-generated content, or multimodal inputs, all of which are common in real-world applications. 
Moreover, the models tested range from 139M to 783M parameters; larger-scale models or instruction-tuned variants might alter the balance observed here. 
Together, these caveats underline that while fine-tuning appears more reliable for simplification, the relative merits of prompting versus fine-tuning remain sensitive to domain, scale, and evaluation perspective.

\section{Ethical Considerations}
\paragraph{Dataset Licensing and Consent}
All models and datasets were used under their official open-source licenses: 
BART, T5, Flan-T5, and Pegasus (Apache 2.0); ProphetNet (MIT); 
WikiLarge/Wikipedia and ASSET (CC-BY-SA); 
Med-EASi (MIT); OneStopEnglish (CC BY-NC-SA). 
Only derivative checkpoints and \textit{WikiLarge-Clean} dataset were released on Hugging Face, 
retaining the original licenses.

\paragraph{Human Evaluation Protocols}
The human study was conducted via anonymous online questionnaires. 
The participants gave their informed consent; no personal data was collected. 
The task was low-risk and limited to text quality judgments.

\paragraph{Bias and Fairness}
Simplification datasets and pretrained models may inherit stylistic or demographic biases. 
Several corpora and model cards already note such limitations. 
We did not conduct a dedicated bias audit but acknowledge this as an open consideration for future work.

\paragraph{Safety and Misuse}
Simplification can distort sensitive material (e.g., medical or legal text), risking user harm. 
Guardrails and disclaimers are recommended for high-stakes deployments.

\paragraph{Environmental Impact}
As described in Section~\ref{sec:ft-setup} for fine-tuning, our dataset cleaning and evaluation were likewise run on Google Cloud managed GPUs (typically a single NVIDIA L4). 
Since execution was abstracted by the provider we cannot precisely estimate the footprint, but overall compute was modest and the environmental impact negligible relative to large-scale pretraining.

\paragraph{Reproducibility and Release}
\label{sec:release}
We release all code, prompts, and configuration files on \footurl{GitHub}{https://github.com/eilamc14/Simplify-This}, together with fine-tuned models checkpoints \footurl{(<model\_name>-text-simplification)}{https://huggingface.co/eilamc14/models} and the \footurl{WikiLarge-Clean dataset}{https://huggingface.co/datasets/eilamc14/wikilarge-clean} on Hugging Face. 
All artifacts retain their original licenses and include model cards for responsible use.

\section{Conclusion and Future Work}
In this work, we systematically compared fine-tuning and prompt engineering for text simplification across multiple datasets and models. 
Our findings showed that fine-tuning remains the most reliable approach for achieving genuine simplification quality, while prompting can sometimes optimize surface-level similarity metrics but is prone to copying.

For future research, several promising directions emerge. 
First, a hybrid strategy of \textbf{lightweight fine-tuning combined with prompt engineering} deserves further study. 
Models such as T5, which struggled under pure prompting in our experiments, may benefit from limited task-specific adaptation that avoids the full cost of large-scale fine-tuning.
Second, evaluation should be extended to \textbf{newer and more diverse models}, including decoder-only architectures and larger LLMs; evidence (e.g., BLESS: \textit{GPT-3.5-Turbo} outperforming \textit{MUSS})\citep{kew-etal-2023-bless} suggests that scaling and training paradigms may alter the relative advantage of fine-tuning versus prompting.
Third, extending the comparison beyond text simplification to \textbf{other NLP tasks and potentially multimodal settings} could reveal whether the observed trade-offs generalize more broadly.
Finally, it is important to situate our findings in the context of \textbf{state-of-the-art frontier models} such as GPT-5, Claude, Grok or Gemini. 
These systems, trained with massive compute budgets and billions of parameters, may rival or even surpass smaller fine-tuned models on simplification benchmarks. 
Yet, given their high cost and limited controllability, it remains an open question whether their advantages translate into consistent improvements on both simple and domain-specific tasks, or whether smaller, fine-tuned models will remain the more efficient and practical choice. 
This tension between \textbf{scalability, cost-efficiency, and task specialization} is central to the next phase of research.

\balance

\bibliography{references}

@inproceedings{brown2020language,
  title     = {Language Models are Few-Shot Learners},
  author    = {Brown, Tom B. and Mann, Benjamin and Ryder, Nick and Subbiah, Melanie and Kaplan, Jared and Dhariwal, Prafulla and Neelakantan, Arvind and Shyam, Pranav and Sastry, Girish and Askell, Amanda and others},
  booktitle = {Proceedings of NeurIPS},
  volume    = {33},
  pages     = {1877--1901},
  year      = {2020}
}

@article{wei2022chain,
  title   = {Chain-of-thought prompting elicits reasoning in large language models},
  author  = {Wei, Jason and Wang, Xuezhi and Schuurmans, Dale and Bosma, Maarten and others},
  journal = {arXiv preprint arXiv:2201.11903},
  year    = {2022}
}

@inproceedings{paetzold2016coling,
  title     = {Understanding the Lexical Simplification Needs of Non-Native Speakers of English},
  author    = {Paetzold, Gustavo Henrique and Specia, Lucia},
  booktitle = {Proceedings of COLING},
  pages     = {717--727},
  year      = {2016}
}

@inproceedings{paetzold2016naacl,
  title     = {Inferring Psycholinguistic Properties of Words},
  author    = {Paetzold, Gustavo Henrique and Specia, Lucia},
  booktitle = {Proceedings of NAACL-HLT},
  pages     = {435--440},
  year      = {2016}
}

@article{siddharthan2014survey,
  title     = {A survey of research on text simplification},
  author    = {Siddharthan, Advaith},
  journal   = {ITL International Journal of Applied Linguistics},
  volume    = {165},
  number    = {2},
  pages     = {259--298},
  year      = {2014},
  publisher = {John Benjamins}
}

@article{alva2020survey,
  title     = {Data-driven sentence simplification: Survey and benchmark},
  author    = {Alva-Manchego, Fernando and Scarton, Carolina and Specia, Lucia},
  journal   = {Computational Linguistics},
  volume    = {46},
  number    = {1},
  pages     = {135--187},
  year      = {2020},
  publisher = {MIT Press}
}

@article{flesch1948yardstick,
  title     = {A new readability yardstick},
  author    = {Flesch, Rudolf},
  journal   = {Journal of Applied Psychology},
  volume    = {32},
  number    = {3},
  pages     = {221--233},
  year      = {1948},
  publisher = {American Psychological Association}
}

@techreport{kincaid1975derivation,
  author = {Kincaid, J. Peter and Fishburne, R. P. and Rogers, R. L. and Chissom, B. S.},
  title  = {Derivation of new readability formulas (Automated Readability Index, Fog Count and Flesch Reading Ease Formula) for Navy enlisted personnel},
  institution = {Research Branch, Chief of Naval Technical Training, U. S. Naval Air Station},
  year   = {1975},
  number = {Research Branch Report 8-75},
}

@article{crossley2014simple,
  title     = {What's so simple about simplified texts? A computational and psycholinguistic investigation of text comprehension and text processing},
  author    = {Crossley, Scott A. and Yang, Hae Sung and McNamara, Danielle S.},
  journal   = {Reading in a Foreign Language},
  volume    = {26},
  number    = {1},
  pages     = {92--113},
  year      = {2014}
}

@article{Shardlow2014,
    title = {A Survey of Automated Text Simplification},
    journal = {International Journal of Advanced Computer Science and Applications(IJACSA), Special Issue on Natural Language Processing 2014},
    doi = {10.14569/SpecialIssue.2014.040109},
    url = {http://dx.doi.org/10.14569/SpecialIssue.2014.040109},
    year = {2014},
    publisher = {The Science and Information Organization},
    volume = {4},
    number = {1},
    author = {Matthew Shardlow}
}

@article{Xu2016OptimizingSM,
  title={Optimizing Statistical Machine Translation for Text Simplification},
  author={Wei Xu and Courtney Napoles and Ellie Pavlick and Quan Ze Chen and Chris Callison-Burch},
  journal={Transactions of the Association for Computational Linguistics},
  year={2016},
  volume={4},
  pages={401-415},
  url={https://api.semanticscholar.org/CorpusID:2177849}
}

@article{DBLP:journals/corr/abs-1904-09675,
  author       = {Tianyi Zhang and
                  Varsha Kishore and
                  Felix Wu and
                  Kilian Q. Weinberger and
                  Yoav Artzi},
  title        = {BERTScore: Evaluating Text Generation with {BERT}},
  journal      = {CoRR},
  volume       = {abs/1904.09675},
  year         = {2019},
  url          = {http://arxiv.org/abs/1904.09675},
  eprinttype    = {arXiv},
  eprint       = {1904.09675},
  bibsource    = {dblp computer science bibliography, https://dblp.org}
}

@article{2020t5,
  author  = {Colin Raffel and Noam Shazeer and Adam Roberts and Katherine Lee and Sharan Narang and Michael Matena and Yanqi Zhou and Wei Li and Peter J. Liu},
  title   = {Exploring the Limits of Transfer Learning with a Unified Text-to-Text Transformer},
  journal = {Journal of Machine Learning Research},
  year    = {2020},
  volume  = {21},
  number  = {140},
  pages   = {1-67},
  url     = {http://jmlr.org/papers/v21/20-074.html}
}

@inproceedings{lewis-etal-2020-bart,
    title = "{BART}: Denoising Sequence-to-Sequence Pre-training for Natural Language Generation, Translation, and Comprehension",
    author = "Lewis, Mike  and
      Liu, Yinhan  and
      Goyal, Naman  and
      Ghazvininejad, Marjan  and
      Mohamed, Abdelrahman  and
      Levy, Omer  and
      Stoyanov, Veselin  and
      Zettlemoyer, Luke",
    editor = "Jurafsky, Dan  and
      Chai, Joyce  and
      Schluter, Natalie  and
      Tetreault, Joel",
    booktitle = "Proceedings of the 58th Annual Meeting of the Association for Computational Linguistics",
    month = jul,
    year = "2020",
    address = "Online",
    publisher = "Association for Computational Linguistics",
    url = "https://aclanthology.org/2020.acl-main.703/",
    doi = "10.18653/v1/2020.acl-main.703",
    pages = "7871--7880"
}

@inproceedings{alva-manchego-etal-2020-asset,
    title = "{ASSET}: {A} Dataset for Tuning and Evaluation of Sentence Simplification Models with Multiple Rewriting Transformations",
    author = "Alva-Manchego, Fernando  and
      Martin, Louis  and
      Bordes, Antoine  and
      Scarton, Carolina  and
      Sagot, Beno{\^i}t  and
      Specia, Lucia",
    editor = "Jurafsky, Dan  and
      Chai, Joyce  and
      Schluter, Natalie  and
      Tetreault, Joel",
    booktitle = "Proceedings of the 58th Annual Meeting of the Association for Computational Linguistics",
    month = jul,
    year = "2020",
    address = "Online",
    publisher = "Association for Computational Linguistics",
    url = "https://aclanthology.org/2020.acl-main.424/",
    doi = "10.18653/v1/2020.acl-main.424",
    pages = "4668--4679"
}

@article{basu2023med,
  title={Med-EASi: Finely Annotated Dataset and Models for Controllable Simplification of Medical Texts},
  author={Basu, Chandrayee and Vasu, Rosni and Yasunaga, Michihiro and Yang, Qian},
  journal={arXiv preprint arXiv:2302.09155},
  year={2023}
}

@inproceedings{kew-etal-2023-bless,
    title = "{BLESS}: Benchmarking Large Language Models on Sentence Simplification",
    author = "Kew, Tannon  and
      Chi, Alison  and
      V{\'a}squez-Rodr{\'i}guez, Laura  and
      Agrawal, Sweta  and
      Aumiller, Dennis  and
      Alva-Manchego, Fernando  and
      Shardlow, Matthew",
    editor = "Bouamor, Houda  and
      Pino, Juan  and
      Bali, Kalika",
    booktitle = "Proceedings of the 2023 Conference on Empirical Methods in Natural Language Processing",
    month = dec,
    year = "2023",
    address = "Singapore",
    publisher = "Association for Computational Linguistics",
    url = "https://aclanthology.org/2023.emnlp-main.821/",
    doi = "10.18653/v1/2023.emnlp-main.821",
    pages = "13291--13309"
}

@misc{chung2022scalinginstructionfinetunedlanguagemodels,
      title={Scaling Instruction-Finetuned Language Models}, 
      author={Hyung Won Chung and Le Hou and Shayne Longpre and Barret Zoph and Yi Tay and William Fedus and Yunxuan Li and Xuezhi Wang and Mostafa Dehghani and Siddhartha Brahma and Albert Webson and Shixiang Shane Gu and Zhuyun Dai and Mirac Suzgun and Xinyun Chen and Aakanksha Chowdhery and Alex Castro-Ros and Marie Pellat and Kevin Robinson and Dasha Valter and Sharan Narang and Gaurav Mishra and Adams Yu and Vincent Zhao and Yanping Huang and Andrew Dai and Hongkun Yu and Slav Petrov and Ed H. Chi and Jeff Dean and Jacob Devlin and Adam Roberts and Denny Zhou and Quoc V. Le and Jason Wei},
      year={2022},
      eprint={2210.11416},
      archivePrefix={arXiv},
      primaryClass={cs.LG},
      url={https://arxiv.org/abs/2210.11416}, 
}

@inproceedings{vajjala-lucic-2018-onestopenglish,
    title = "{O}ne{S}top{E}nglish corpus: A new corpus for automatic readability assessment and text simplification",
    author = "Vajjala, Sowmya  and
      Lu{\v{c}}i{\'c}, Ivana",
    editor = "Tetreault, Joel  and
      Burstein, Jill  and
      Kochmar, Ekaterina  and
      Leacock, Claudia  and
      Yannakoudakis, Helen",
    booktitle = "Proceedings of the Thirteenth Workshop on Innovative Use of {NLP} for Building Educational Applications",
    month = jun,
    year = "2018",
    address = "New Orleans, Louisiana",
    publisher = "Association for Computational Linguistics",
    url = "https://aclanthology.org/W18-0535/",
    doi = "10.18653/v1/W18-0535",
    pages = "297--304"
}

@misc{bahdanau2016neuralmachinetranslationjointly,
      title={Neural Machine Translation by Jointly Learning to Align and Translate}, 
      author={Dzmitry Bahdanau and Kyunghyun Cho and Yoshua Bengio},
      year={2016},
      eprint={1409.0473},
      archivePrefix={arXiv},
      primaryClass={cs.CL},
      url={https://arxiv.org/abs/1409.0473}, 
}

@inproceedings{see-etal-2017-get,
    title = "Get To The Point: Summarization with Pointer-Generator Networks",
    author = "See, Abigail  and
      Liu, Peter J.  and
      Manning, Christopher D.",
    editor = "Barzilay, Regina  and
      Kan, Min-Yen",
    booktitle = "Proceedings of the 55th Annual Meeting of the Association for Computational Linguistics (Volume 1: Long Papers)",
    month = jul,
    year = "2017",
    address = "Vancouver, Canada",
    publisher = "Association for Computational Linguistics",
    url = "https://aclanthology.org/P17-1099/",
    doi = "10.18653/v1/P17-1099",
    pages = "1073--1083"
}

@misc{vinyals2015neuralconversationalmodel,
      title={A Neural Conversational Model}, 
      author={Oriol Vinyals and Quoc Le},
      year={2015},
      eprint={1506.05869},
      archivePrefix={arXiv},
      primaryClass={cs.CL},
      url={https://arxiv.org/abs/1506.05869}, 
}

@inproceedings{zhang-lapata-2017-sentence,
    title = "Sentence Simplification with Deep Reinforcement Learning",
    author = "Zhang, Xingxing  and
      Lapata, Mirella",
    editor = "Palmer, Martha  and
      Hwa, Rebecca  and
      Riedel, Sebastian",
    booktitle = "Proceedings of the 2017 Conference on Empirical Methods in Natural Language Processing",
    month = sep,
    year = "2017",
    address = "Copenhagen, Denmark",
    publisher = "Association for Computational Linguistics",
    url = "https://aclanthology.org/D17-1062/",
    doi = "10.18653/v1/D17-1062",
    pages = "584--594"
}

@misc{martin2021mussmultilingualunsupervisedsentence,
      title={MUSS: Multilingual Unsupervised Sentence Simplification by Mining Paraphrases}, 
      author={Louis Martin and Angela Fan and Éric de la Clergerie and Antoine Bordes and Benoît Sagot},
      year={2021},
      eprint={2005.00352},
      archivePrefix={arXiv},
      primaryClass={cs.CL},
      url={https://arxiv.org/abs/2005.00352}, 
}

@misc{wolf2020huggingfacestransformersstateoftheartnatural,
      title={HuggingFace's Transformers: State-of-the-art Natural Language Processing}, 
      author={Thomas Wolf and Lysandre Debut and Victor Sanh and Julien Chaumond and Clement Delangue and Anthony Moi and Pierric Cistac and Tim Rault and Rémi Louf and Morgan Funtowicz and Joe Davison and Sam Shleifer and Patrick von Platen and Clara Ma and Yacine Jernite and Julien Plu and Canwen Xu and Teven Le Scao and Sylvain Gugger and Mariama Drame and Quentin Lhoest and Alexander M. Rush},
      year={2020},
      eprint={1910.03771},
      archivePrefix={arXiv},
      primaryClass={cs.CL},
      url={https://arxiv.org/abs/1910.03771}, 
}

@misc{zhang2020pegasuspretrainingextractedgapsentences,
      title={PEGASUS: Pre-training with Extracted Gap-sentences for Abstractive Summarization}, 
      author={Jingqing Zhang and Yao Zhao and Mohammad Saleh and Peter J. Liu},
      year={2020},
      eprint={1912.08777},
      archivePrefix={arXiv},
      primaryClass={cs.CL},
      url={https://arxiv.org/abs/1912.08777}, 
}

@inproceedings{narayan-etal-2018-dont,
    title = "Don{'}t Give Me the Details, Just the Summary! Topic-Aware Convolutional Neural Networks for Extreme Summarization",
    author = "Narayan, Shashi  and
      Cohen, Shay B.  and
      Lapata, Mirella",
    editor = "Riloff, Ellen  and
      Chiang, David  and
      Hockenmaier, Julia  and
      Tsujii, Jun{'}ichi",
    booktitle = "Proceedings of the 2018 Conference on Empirical Methods in Natural Language Processing",
    month = oct # "-" # nov,
    year = "2018",
    address = "Brussels, Belgium",
    publisher = "Association for Computational Linguistics",
    url = "https://aclanthology.org/D18-1206/",
    doi = "10.18653/v1/D18-1206",
    pages = "1797--1807"
}

@misc{qi2020prophetnetpredictingfuturengram,
      title={ProphetNet: Predicting Future N-gram for Sequence-to-Sequence Pre-training}, 
      author={Weizhen Qi and Yu Yan and Yeyun Gong and Dayiheng Liu and Nan Duan and Jiusheng Chen and Ruofei Zhang and Ming Zhou},
      year={2020},
      eprint={2001.04063},
      archivePrefix={arXiv},
      primaryClass={cs.CL},
      url={https://arxiv.org/abs/2001.04063}, 
}

@misc{hermann2015teachingmachinesreadcomprehend,
      title={Teaching Machines to Read and Comprehend}, 
      author={Karl Moritz Hermann and Tomáš Kočiský and Edward Grefenstette and Lasse Espeholt and Will Kay and Mustafa Suleyman and Phil Blunsom},
      year={2015},
      eprint={1506.03340},
      archivePrefix={arXiv},
      primaryClass={cs.CL},
      url={https://arxiv.org/abs/1506.03340}, 
}

@misc{maddela2023lenslearnableevaluationmetric,
      title={LENS: A Learnable Evaluation Metric for Text Simplification}, 
      author={Mounica Maddela and Yao Dou and David Heineman and Wei Xu},
      year={2023},
      eprint={2212.09739},
      archivePrefix={arXiv},
      primaryClass={cs.CL},
      url={https://arxiv.org/abs/2212.09739}, 
}

@misc{heineman2023dancingsuccessfailureeditlevel,
      title={Dancing Between Success and Failure: Edit-level Simplification Evaluation using SALSA}, 
      author={David Heineman and Yao Dou and Mounica Maddela and Wei Xu},
      year={2023},
      eprint={2305.14458},
      archivePrefix={arXiv},
      primaryClass={cs.CL},
      url={https://arxiv.org/abs/2305.14458}, 
}

@inproceedings{von-werra-etal-2022-evaluate,
    title = "Evaluate {\&} Evaluation on the Hub: Better Best Practices for Data and Model Measurements",
    author = "Von Werra, Leandro  and
      Tunstall, Lewis  and
      Thakur, Abhishek  and
      Luccioni, Sasha  and
      Thrush, Tristan  and
      Piktus, Aleksandra  and
      Marty, Felix  and
      Rajani, Nazneen  and
      Mustar, Victor  and
      Ngo, Helen",
    editor = "Che, Wanxiang  and
      Shutova, Ekaterina",
    booktitle = "Proceedings of the 2022 Conference on Empirical Methods in Natural Language Processing: System Demonstrations",
    month = dec,
    year = "2022",
    address = "Abu Dhabi, UAE",
    publisher = "Association for Computational Linguistics",
    url = "https://aclanthology.org/2022.emnlp-demos.13/",
    doi = "10.18653/v1/2022.emnlp-demos.13",
    pages = "128--136"
}

@inproceedings{pang-gimpel-2019-unsupervised,
    title = "Unsupervised Evaluation Metrics and Learning Criteria for Non-Parallel Textual Transfer",
    author = "Pang, Richard Yuanzhe  and
      Gimpel, Kevin",
    editor = "Birch, Alexandra  and
      Finch, Andrew  and
      Hayashi, Hiroaki  and
      Konstas, Ioannis  and
      Luong, Thang  and
      Neubig, Graham  and
      Oda, Yusuke  and
      Sudoh, Katsuhito",
    booktitle = "Proceedings of the 3rd Workshop on Neural Generation and Translation",
    month = nov,
    year = "2019",
    address = "Hong Kong",
    publisher = "Association for Computational Linguistics",
    url = "https://aclanthology.org/D19-5614/",
    doi = "10.18653/v1/D19-5614",
    pages = "138--147"
}

@misc{krishna2020reformulatingunsupervisedstyletransfer,
      title={Reformulating Unsupervised Style Transfer as Paraphrase Generation}, 
      author={Kalpesh Krishna and John Wieting and Mohit Iyyer},
      year={2020},
      eprint={2010.05700},
      archivePrefix={arXiv},
      primaryClass={cs.CL},
      url={https://arxiv.org/abs/2010.05700}, 
}

@inproceedings{alva-manchego-etal-2019-easse,
    title = "{EASSE}: Easier Automatic Sentence Simplification Evaluation",
    author = "Alva-Manchego, Fernando  and
      Martin, Louis  and
      Scarton, Carolina  and
      Specia, Lucia",
    editor = "Pad{\'o}, Sebastian  and
      Huang, Ruihong",
    booktitle = "Proceedings of the 2019 Conference on Empirical Methods in Natural Language Processing and the 9th International Joint Conference on Natural Language Processing (EMNLP-IJCNLP): System Demonstrations",
    month = nov,
    year = "2019",
    address = "Hong Kong, China",
    publisher = "Association for Computational Linguistics",
    url = "https://aclanthology.org/D19-3009/",
    doi = "10.18653/v1/D19-3009",
    pages = "49--54"
}

@article{WuArase2025,
  author    = {Xuanxin Wu and Yuki Arase},
  title     = {An In-depth Evaluation of Large Language Models in Sentence Simplification with Error-based Human Assessment},
  journal   = {ACM Transactions on Intelligent Systems and Technology},
  year      = {2025},
  publisher = {Association for Computing Machinery},
  doi       = {10.1145/3744744},
  url       = {https://dl.acm.org/doi/10.1145/3744744},
  note      = {Also available as arXiv:2403.04963}
}

@inproceedings{to-etal-2024-deakinnlp,
  title     = {DeakinNLP at BioLaySumm: Evaluating Fine-tuning Longformer and {GPT}-4 Prompting for Biomedical Lay Summarization},
  author    = {To, Huy Quoc and Liu, Ming and Huang, Guangyan},
  booktitle = {Proceedings of the 23rd Workshop on Biomedical Natural Language Processing},
  address   = {Bangkok, Thailand},
  publisher = {Association for Computational Linguistics},
  year      = {2024},
  pages     = {748--754},
  doi       = {10.18653/v1/2024.bionlp-1.67},
  url       = {https://aclanthology.org/2024.bionlp-1.67/}
}

@misc{Fang2025ProgDS,
  author    = {Dengzhao Fang and Jipeng Qiang and Yi Zhu and Yunhao Yuan and Wei Li and Yan Liu},
  title     = {Progressive Document-level Text Simplification via Large Language Models},
  year      = {2025},
  eprint    = {2501.03857},
  archivePrefix = {arXiv},
  primaryClass  = {cs.CL},
  url       = {https://arxiv.org/abs/2501.03857}
}

@inproceedings{farajidizaji-etal-2024-possible,
  title     = {Is It Possible to Modify Text to a Target Readability Level? An Initial Investigation Using Zero-Shot Large Language Models},
  author    = {Farajidizaji, Asma and Raina, Vatsal and Gales, Mark},
  booktitle = {Proceedings of the 2024 Joint International Conference on Computational Linguistics, Language Resources and Evaluation (LREC-COLING 2024)},
  address   = {Torino, Italia},
  publisher = {ELRA and ICCL},
  year      = {2024},
  pages     = {9325--9339},
  url       = {https://aclanthology.org/2024.lrec-main.815/}
}

@article{Xu2015Newsela,
  title={Problems in Current Text Simplification Research: New Data Can Help},
  author={Xu, Wei and Callison-Burch, Chris and Napoles, Courtney},
  journal={Transactions of the Association for Computational Linguistics},
  volume={3},
  pages={283--297},
  year={2015},
  publisher={MIT Press},
  doi={10.1162/tacl_a_00139}
}
\bibliographystyle{plainnat}

\clearpage
\appendix
\section*{Appendix}

\section{WikiLarge Dataset Cleaning}
\label{sec:appendix-dataset}
The cleaning procedure followed the order defined in our preprocessing code:

\begin{itemize}
    \item \textbf{Length constraints.} We discarded pairs with extreme lengths to keep examples within a reasonable complexity band for simplification. Concretely, source sentences were required to have $4 \leq \text{tokens} \leq 256$, and target sentences $2 \leq \text{tokens} \leq 256$ (measured via whitespace tokenization).
    \item \textbf{Compression ratio.} To avoid pathological over-/under-simplification, we retained pairs only if the simple/complex token ratio satisfied $0.40 \leq \mathrm{CR} \leq 0.95$.
    \item \textbf{Near-identity (lexical Jaccard similarity) filtering.} To reduce trivial copying, we removed pairs whose lexical Jaccard similarity between source and target exceeded $0.98$.
    \item \textbf{Deduplication.} After text normalization (lowercasing and whitespace collapsing), we removed duplicate sentence pairs, retaining only the first occurrence. Full implementation details of our stable deduplication key are provided in the dataset card and Appendix.
\end{itemize}

\begin{center}
\begin{minipage}{\linewidth}
\centering
\begin{tabular}{lcc}
\toprule
 & Count & \% \\
\midrule
Initial size & 296{,}402 & -- \\
Final size   & 123{,}862 & -- \\
Length fails & 15{,}764 & 5.3 \\
Compression fails & 156{,}027 & 57.8 \\
Jaccard similarity fails & 633 & 0.9 \\
Deduplication & 116 & 0.04 \\
\bottomrule
\end{tabular}
\captionof{table}{Cleaning statistics for the train split of WikiLarge-Clean.}
\label{tab:cleaning-train1}
\end{minipage}

\end{center}

Table~\ref{tab:cleaning-train1} reports the filtering statistics applied to the WikiLarge training split. 
Out of nearly 300k sentence pairs, more than half were removed due to excessive compression (57.8\%), with additional reductions caused by overly short or long sequences (5.3\%), near-identity pairs (0.9\%), and deduplication. 
The resulting WikiLarge-Clean corpus contains roughly 124k high-quality pairs. 

\section{Fine-Tune Setup}
\label{sec:appendix-finetune}

\paragraph{Hyperparameters.}
Typical default configuration: \textbf{epochs} 5, \textbf{learning rate} $\sim$$3 \times 10^{-5}$, \textbf{optimizer}~AdamW, \textbf{precision}~bfloat16, \textbf{weight decay}~0.01, \textbf{label smoothing}~0.1, \textbf{warmup ratio}~0.1, and \textbf{max gradient norm}~0.5. checkpoint saving and evaluation were scheduled at regular intervals proportional to the effective batch size, and gradient checkpointing was used where necessary to fit large models.

\paragraph{On-the-fly Evaluation and Decoding}
During validation we decode with a fixed generation profile to ensure comparability: \texttt{max\_new\_tokens=64}, \texttt{num\_beams=4}, \texttt{length\_penalty=1.0}, \texttt{no\_repeat\_ngram\_size=3}, \texttt{early\_stopping=True}, \texttt{do\_sample=False}. After removing the prefix \texttt{``Simplify: ''} from the sources we report SARI from 'evaluate' by Hugging Face \citep{von-werra-etal-2022-evaluate} and not-normalized identical ratio for sanity check.

\section{Evaluation Metrics}
\label{sec:appendix-metrics}
We used the \texttt{EASSE} toolkit \citep{alva-manchego-etal-2019-easse} to compute \textbf{SARI}, \textbf{FKGL}, and \textbf{BERTScore}.

\paragraph{SARI}  
SARI \citep{Xu2016OptimizingSM} evaluates simplification by comparing system outputs against both the source and (optional) multiple human references. It rewards appropriate additions, deletions, and keep operations, thus directly modeling the simplification process. SARI has become the de facto standard for automatic evaluation of text simplification, especially when multiple references are available (e.g., ASSET).

\paragraph{FKGL}  
The Flesch–Kincaid Grade Level \citep{flesch1948yardstick,kincaid1975derivation} is a readability metric based on sentence length and syllable counts. While not specific to simplification, it provides an interpretable measure of linguistic complexity, mapping to U.S. school grade levels. Lower FKGL values indicate simpler text. Despite its simplicity, FKGL is frequently reported as a complementary metric in simplification research.

\paragraph{BERTScore}  
BERTScore \citep{DBLP:journals/corr/abs-1904-09675} computes similarity between candidate and reference texts using contextual embeddings from pre-trained language models. It captures semantic adequacy and meaning preservation more effectively than n-gram overlap metrics. We report the F1 variant, following common practice in simplification tasks.

\paragraph{LENS}  
LENS \citep{maddela2023lenslearnableevaluationmetric} is a recent evaluation framework for simplification that integrates language models into a learned metric. It has shown strong correlation with human judgments across multiple datasets, and complements metrics like SARI by covering settings with limited or no references through its \textit{LENS-SALSA} variant.

\paragraph{Identical Ratio} 
To complement existing metrics, we compute the \textbf{Identical Ratio (id\_ratio)}, defined as the fraction of system outputs that are identical to the input after normalization (strip, Unicode NFKC, whitespace collapsing) of both. Formally, we report case-sensitive for most models and a case-insensitive variant for uncased models. This diagnostic is useful because text simplification systems can degenerate into copying the input without applying any meaningful edits, a failure mode noted in prior analyses of simplification evaluation \citep{alva2020survey}. Traditional semantic similarity metrics such as BERTScore are known to be biased toward conservative systems that perform few or no edits \citep{pang-gimpel-2019-unsupervised,krishna2020reformulatingunsupervisedstyletransfer}, while SARI penalizes copying but can still mask trivial behavior depending on reference coverage \citep{Xu2016OptimizingSM}. Reporting the identical ratio thus provides a transparent diagnostic: a high id\_ratio indicates that a model achieves scores primarily by avoiding edits, whereas a lower value suggests a more substantive simplification. $id\_ratio=0$ can sometimes indicate that the model constantly repeats parts of the prompt or gives unrelated outputs. This measure is not intended as a standalone quality metric but rather as a sanity check and complement to SARI, FKGL, BERTScore, and LENS.

\section{Prompt Templates}
\label{sec:appendix-prompts}
We implemented ten prompt templates (P1--P10) to elicit simplification behavior from untuned models. The designs cover major prompting strategies: zero-shot instruction \citep{brown2020language}, multi-shot demonstrations, chain-of-thought reasoning \citep{wei2022chain}, lexical and psycholinguistic constraints \citep{paetzold2016coling, paetzold2016naacl}, sentence splitting \citep{siddharthan2014survey}, data-driven transformation cues \citep{alva2020survey}, readability control \citep{flesch1948yardstick}, ESL comprehension \citep{crossley2014simple}, and content-preservation constraints \citep{alva2020survey}. We added P0 - control prompt which is identical to the preprocessing the fine-tune models were trained on.

Each template required the model to output only the simplified sentence, with the source inserted directly into the instruction. Multi-shot prompts used a fixed set of three examples across models. All prompt-based runs employed the same decoding settings as fine-tuned models. The complete prompts and their theoretical motivations are provided in Table~\ref{tab:prompts}.

\begin{table*}[t]
\centering
\caption{Prompt templates (P0--P10) for text simplification. <src> represents the source sentence inserted to the prompt.}
\label{tab:prompts}
\begin{tabularx}{\textwidth}{@{} l p{2.2cm} Y p{1.8cm} @{}}
\toprule
\specialrule{0.1pt}{2pt}{2pt}
ID & Strategy & Prompt & Reference \\
\specialrule{0.1pt}{2pt}{2pt}
\toprule
P0  & Control                   & Simplify: <src> & --- \\ \specialrule{0.1pt}{2pt}{2pt}
P1  & Zero-shot instruction     & Simplify the following sentence so it is easy to understand while keeping the original meaning: <src> & \citep{brown2020language} \\ \specialrule{0.1pt}{2pt}{2pt}
P2  & Multi-shot examples       & Simplify the sentence. Use common words; keep the meaning. Output only the simplified sentence.
Complex: The committee reached a unanimous decision after extensive deliberations. Simple: The group agreed after talking for a long time.
Complex: The ancient manuscript was preserved in a climate-controlled archive to prevent deterioration. Simple: The old book was kept in a special room to stop it from getting damaged.
Complex: The economic downturn had a profound effect on small businesses across the region. Simple: The bad economy hurt many small businesses in the area.
Complex: <src> Simple: & \citep{brown2020language} \\ \specialrule{0.1pt}{2pt}{2pt}
P3  & Chain-of-thought          & First list the words/phrases that make this sentence hard to read. Then rewrite the sentence in simpler language without changing its meaning: <src> & \citep{wei2022chain} \\ \specialrule{0.1pt}{2pt}{2pt}
P4  & Lexical simplification (L2) & Rewrite the sentence using common, high-frequency words suitable for a B1 (intermediate) non-native reader. Keep all original information: <src> & \citep{paetzold2016coling} \\ \specialrule{0.1pt}{2pt}{2pt}
P5  & Psycholinguistic constraints & Rewrite the sentence using words with high familiarity and early age-of-acquisition (avoid abstract or rare terms). Preserve the original meaning: <src> & \citep{paetzold2016naacl} \\ \specialrule{0.1pt}{2pt}{2pt}
P6  & Sentence splitting        & Rewrite the sentence in simpler words and split long or embedded clauses into shorter sentences. Keep the same meaning: <src> & \citep{siddharthan2014survey} \\ \specialrule{0.1pt}{2pt}{2pt}
P7  & Transformation cues       & Apply common simplification transformations (e.g., replace complex words, reorder for clarity, split long clauses) while keeping grammar and meaning: <src> & \citep{alva2020survey} \\ \specialrule{0.1pt}{2pt}{2pt}
P8  & Readability target        & Rewrite the sentence so that it reaches a Flesch Reading Ease score $\geq 80$ ($\approx$ grade 6), without losing information: <src> & \citep{flesch1948yardstick} \\ \specialrule{0.1pt}{2pt}{2pt}
P9  & ESL comprehension         & Rewrite the sentence for ESL learners - use high-frequency words, avoid idioms, and add brief clarifications if needed. Keep the meaning the same: <src> & \citep{crossley2014simple} \\ \specialrule{0.1pt}{2pt}{2pt}
P10 & Content preservation      & Simplify the sentence for readability, but preserve ALL factual details (entities, quantities, relations) exactly: <src> & \citep{alva2020survey} \\ \specialrule{0.1pt}{2pt}{2pt}
\bottomrule
\end{tabularx}
\end{table*}

\onecolumn

\section{Full Results Tables}
\label{app:full_tables}
Tables ~\ref{tab:asset-all-models-full}, ~\ref{tab:medeasi-all-models-full} and ~\ref{tab:ose-all-models-full} for ASSET, Med-EASi and OneStopEnglish respectively show full per-prompt results across all models. Legend: \textbf{FT} = Fine-tuned; \textbf{PB} = Prompt-based; 
Case-sensitive Identical ratio unless added \textbf{(ci)}; rows with ratio $>0.50$ marked with $\dagger$.

\begin{longtable}{l l c c c c c c}
\caption{ASSET - Full per-prompt results\label{tab:asset-all-models-full}}\\
\toprule
Model & Variant & Prompt\# & Identical ratio & SARI $\uparrow$ & FKGL $\downarrow$ & BERTScore $\uparrow$ & LENS $\uparrow$ \\
\midrule
\endfirsthead
\toprule
Model & Variant & Prompt\# & Identical ratio & SARI $\uparrow$ & FKGL $\downarrow$ & BERTScore $\uparrow$ & LENS $\uparrow$ \\
\midrule
\endhead
\midrule
\multicolumn{7}{r}{\emph{(table continues on next page)}} \\
\midrule
\endfoot
\bottomrule
\endlastfoot

\multirow{12}{*}{BART-base}
  & FT & -- & 0.00 & \textbf{36.13} & 8.50 & 85.57 & 43.88 \\
  & PB & 0  & 0.93$^\dagger$  & 21.44 & 10.02 & \textbf{90.80} & \textbf{60.05} \\
  & PB & 1  & 0.90$^\dagger$  & 22.61 & 9.94  & 90.19 & 58.44 \\
  & PB & 2  & 0.00 & 26.70 & 5.20  & -6.92 & 4.76 \\
  & PB & 3  & 0.77$^\dagger$ & 25.30 & 9.63  & 88.16 & 53.79 \\
  & PB & 4  & 0.73$^\dagger$ & 26.15 & 9.49  & 87.53 & 53.07 \\
  & PB & 5  & 0.73$^\dagger$ & 26.15 & 9.50  & 87.75 & 53.17 \\
  & PB & 6  & 0.85$^\dagger$ & 23.48 & 9.85  & 89.78 & 57.26 \\
  & PB & 7  & 0.75$^\dagger$ & 25.77 & 9.56  & 88.01 & 53.44 \\
  & PB & 8  & 0.77$^\dagger$ & 25.41 & 9.64  & 88.22 & 53.79 \\
  & PB & 9  & 0.70$^\dagger$ & 26.88 & 9.38  & 86.59 & 51.58 \\
  & PB & 10 & 0.84$^\dagger$ & 24.09 & 9.77  & 89.48 & 56.21 \\ 

\midrule

\multirow{12}{*}{BART-large}
  & FT       & --  & 0.01            & \textbf{37.96} & 7.87 & 84.64 & 44.51 \\
  & PB     & 0   & 0.92$^\dagger$  & 21.81 & 9.78  & \textbf{90.38} & \textbf{59.55} \\
  & PB     & 1   & 0.88$^\dagger$  & 23.15 & 9.76  & 89.87 & 57.77 \\
  & PB     & 2   & 0.00            & 31.61 & 4.87  & 30.91 & 12.28 \\
  & PB     & 3   & 0.75$^\dagger$  & 25.67 & 9.42  & 87.52 & 53.62 \\
  & PB     & 4   & 0.71$^\dagger$  & 26.26 & 9.38  & 86.95 & 52.56 \\
  & PB     & 5   & 0.72$^\dagger$  & 26.18 & 9.43  & 87.49 & 53.07 \\
  & PB     & 6   & 0.84$^\dagger$  & 23.79 & 9.69  & 89.37 & 57.05 \\
  & PB     & 7   & 0.72$^\dagger$  & 26.30 & 9.19  & 86.97 & 52.71 \\
  & PB     & 8   & 0.76$^\dagger$  & 25.53 & 9.50  & 87.64 & 53.30 \\
  & PB     & 9   & 0.69$^\dagger$  & 27.14 & 9.29  & 86.31 & 51.48 \\
  & PB   & 10  & 0.83$^\dagger$  & 24.41 & 9.65  & 89.14 & 55.98 \\

\midrule

\multirow{12}{*}{T5-base}
  & FT       & --  & 0.20  & \textbf{35.38} & 8.45 & \textbf{86.85} & \textbf{57.43} \\
  & PB     & 0   & 0.27  & 29.11 & 8.61  & 81.96 & 49.27 \\
  & PB     & 1   & 0.12  & 32.25 & 3.22  & 18.53 & 14.70 \\
  & PB     & 2   & 0.00  & 23.68 & 0.02  & -8.33 & 0.99 \\
  & PB     & 3   & 0.00  & 23.03 & 0.00  & -8.04 & 0.35 \\
  & PB     & 4   & 0.15  & 34.00 & 6.05  & 30.20 & 25.41 \\
  & PB     & 5   & 0.11  & 32.08 & 5.85  & 17.74 & 15.13 \\
  & PB     & 6   & 0.00  & 23.34 & 0.00  & -6.76 & 0.70 \\
  & PB     & 7   & 0.06  & 29.36 & 2.49  & 7.57 & 9.39 \\
  & PB     & 8   & 0.00  & 23.14 & 0.00  & -8.50 & 0.48 \\
  & PB     & 9   & 0.01  & 35.05 & 6.96  & 22.16 & 30.00 \\
  & PB     & 10  & 0.07  & 30.89 & 4.04  & 17.16 & 13.43 \\

\midrule

\multirow{12}{*}{T5-large}
  & FT       & --  & 0.22  & 35.41 & 8.54 & \textbf{87.04} & \textbf{59.81} \\
  & PB     & 0   & 0.01  & 33.86 & 0.17  & 46.29 & 3.99 \\
  & PB     & 1   & 0.00  & 31.86 & 1.63  & 46.60 & 6.46 \\
  & PB     & 2   & 0.00  & \textbf{36.18} & 4.79  & 44.43 & 5.07 \\
  & PB     & 3   & 0.00  & 32.71 & 1.41  & 45.37 & 10.99 \\
  & PB     & 4   & 0.00  & 34.84 & 2.62  & 40.40 & 6.25 \\
  & PB     & 5   & 0.00  & 33.68 & 4.69  & 37.39 & 3.09 \\
  & PB     & 6   & 0.00  & 32.89 & 0.81  & 40.26 & 5.03 \\
  & PB     & 7   & 0.00  & 36.18 & 3.29  & 36.22 & 5.12 \\
  & PB     & 8   & 0.00  & 34.47 & 0.30  & 42.01 & 3.42 \\
  & PB     & 9   & 0.00  & 32.55 & 3.26  & 42.28 & 8.55 \\
  & PB     & 10  & 0.00  & 32.38 & 2.61  & 44.78 & 4.38 \\

\midrule

\multirow{12}{*}{Flan-T5-base}
  & FT   & --  & 0.01            & \textbf{37.92} & 8.22 & 84.40 & 48.29 \\
  & PB & 0   & 0.13            & 36.23 & 6.91 & 58.13 & 41.14 \\
  & PB & 1   & 0.75$^\dagger$  & 24.66 & 9.72 & \textbf{90.17} & 60.62 \\
  & PB & 2   & 0.31            & 31.30 & 9.19 & 87.69 & 62.90 \\
  & PB & 3   & 0.35            & 31.20 & 9.25 & 88.50 & 62.96 \\
  & PB & 4   & 0.31            & 33.37 & 8.62 & 88.67 & \textbf{64.53} \\
  & PB & 5   & 0.47            & 29.24 & 9.43 & 89.32 & 62.17 \\
  & PB & 6   & 0.23            & 34.32 & 8.71 & 88.27 & 64.47 \\
  & PB & 7   & 0.74$^\dagger$  & 24.69 & 9.67 & 89.24 & 60.31 \\
  & PB & 8   & 0.38            & 31.89 & 9.10 & 89.04 & 63.59 \\
  & PB & 9   & 0.38            & 31.20 & 9.13 & 88.97 & 62.98 \\
  & PB & 10  & 0.70$^\dagger$  & 25.63 & 9.66 & 89.59 & 61.17 \\

  \midrule
  
\multirow{12}{*}{Flan-T5-large}
  & FT   & --  & 0.02            & \textbf{37.91} & 7.86 & 84.03 & 47.87 \\
  & PB & 0   & 0.26            & 34.57 & 8.22 & 85.14 & 63.31 \\
  & PB & 1   & 0.40            & 29.51 & 9.42 & 89.41 & 61.35 \\
  & PB & 2   & 0.32            & 32.56 & 8.95 & 89.01 & 63.92 \\
  & PB & 3   & 0.30            & 32.97 & 8.86 & 88.66 & 63.24 \\
  & PB & 4   & 0.22            & 36.31 & 8.28 & 86.79 & \textbf{66.31} \\
  & PB & 5   & 0.40            & 30.76 & 9.24 & 89.52 & 62.88 \\
  & PB & 6   & 0.43            & 30.23 & 9.33 & 90.04 & 62.47 \\
  & PB & 7   & 0.67$^\dagger$  & 26.00 & 9.68 & 89.46 & 60.00 \\
  & PB & 8   & 0.60$^\dagger$  & 27.70 & 9.56 & \textbf{90.09} & 61.60 \\
  & PB & 9   & 0.43            & 31.31 & 9.09 & 89.65 & 63.88 \\
  & PB & 10  & 0.40            & 29.30 & 9.52 & 89.54 & 61.28 \\

\midrule

\multirow{12}{*}{Pegasus-large}
  & FT   & --  & 0.24            & \textbf{35.67} & 8.79 & 86.25 & \textbf{61.52} \\
  & PB & 0   & 0.86$^\dagger$  & 23.33 & 10.24 & \textbf{88.74} & 58.48 \\
  & PB & 1   & 0.84$^\dagger$  & 23.52 & 10.30 & 88.54 & 57.90 \\
  & PB & 2   & 0.00            & 26.37 & 9.64 & 19.66 & 30.97 \\
  & PB & 3   & 0.00            & 23.68 & 14.77 & 58.43 & 38.45 \\
  & PB & 4   & 0.00            & 23.37 & 13.21 & 68.95 & 24.99 \\
  & PB & 5   & 0.00            & 23.81 & 12.75 & 68.46 & 32.32 \\
  & PB & 6   & 0.00            & 23.46 & 11.09 & 67.18 & 22.98 \\
  & PB & 7   & 0.77$^\dagger$  & 25.41 & 9.96  & 87.32 & 54.74 \\
  & PB & 8   & 0.00            & 25.24 & 16.22 & 43.00 & 14.05 \\
  & PB & 9   & 0.00            & 23.12 & 11.29 & 66.94 & 29.90 \\
  & PB & 10  & 0.82$^\dagger$  & 23.99 & 10.21 & 88.11 & 56.72 \\

\midrule

\multirow{12}{*}{Pegasus-xsum}
  & FT   & --  & 0.29            & 33.80 & 9.23 & \textbf{87.54} & 62.46 \\
  & PB & 0   & 0.03            & 32.03 & 8.47 & 29.10 & 52.74 \\
  & PB & 1   & 0.05            & 33.32 & 8.07 & 29.33 & 55.98 \\
  & PB & 2   & 0.00            & 25.12 & 3.11 & 2.76  & 55.24 \\
  & PB & 3   & 0.01            & 28.63 & 5.36 & 12.34 & \textbf{65.49} \\
  & PB & 4   & 0.03            & 30.33 & 6.50 & 16.34 & 57.30 \\
  & PB & 5   & 0.05            & 33.54 & 7.12 & 32.92 & 54.78 \\
  & PB & 6   & 0.00            & 27.43 & 4.48 & 12.74 & 60.25 \\
  & PB & 7   & 0.02            & 30.51 & 6.52 & 21.56 & 63.47 \\
  & PB & 8   & 0.06            & \textbf{34.52} & 8.08 & 37.17 & 57.97 \\
  & PB & 9   & 0.05            & 31.34 & 6.79 & 21.34 & 59.21 \\
  & PB & 10  & 0.02            & 29.81 & 6.76 & 14.81 & 56.68 \\

\midrule

\noalign{\penalty -10000}
\multirow{12}{*}{\shortstack[l]{ProphetNet-large-\\uncased-cnndm}}
  & FT   & --  & 0.11 (ci) & \textbf{38.01} & 7.70 & \textbf{67.82} & \textbf{60.85} \\
  & PB & 0   & 0.16 (ci) & 37.76 & 5.75 & 62.30 & 51.60 \\
  & PB & 1   & 0.22 (ci) & 34.58 & 8.00 & 64.85 & 51.63 \\
  & PB & 2   & 0.00 (ci) & 32.40 & 7.83 & 17.34 & 45.19 \\
  & PB & 3   & 0.06 (ci) & 36.79 & 5.20 & 28.20 & 48.60 \\
  & PB & 4   & 0.04 (ci) & 37.36 & 7.37 & 27.18 & 32.43 \\
  & PB & 5   & 0.03 (ci) & 37.50 & 8.25 & 38.09 & 31.44 \\
  & PB & 6   & 0.08 (ci) & 37.58 & 8.01 & 40.47 & 38.40 \\
  & PB & 7   & 0.13 (ci) & 35.15 & 6.89 & 57.87 & 42.56 \\
  & PB & 8   & 0.13 (ci) & 37.25 & 6.27 & 50.08 & 40.97 \\
  & PB & 9   & 0.05 (ci) & 37.13 & 7.04 & 37.79 & 51.60 \\
  & PB & 10  & 0.12 (ci) & 36.06 & 8.54 & 54.66 & 45.61 \\

\end{longtable}

\begin{longtable}{l l c c c c c c}
\caption{Med-EASi - Full per-prompt results\label{tab:medeasi-all-models-full}}\\
\toprule
Model & Variant & Prompt\# & Identical ratio & SARI $\uparrow$ & FKGL $\downarrow$ & BERTScore $\uparrow$ & LENS $\uparrow$ \\
\midrule
\endfirsthead
\toprule
Model & Variant & Prompt\# & Identical ratio & SARI $\uparrow$ & FKGL $\downarrow$ & BERTScore $\uparrow$ & LENS $\uparrow$ \\
\midrule
\endhead
\midrule
\multicolumn{7}{r}{\emph{(table continues on next page)}} \\
\midrule
\endfoot
\bottomrule
\endlastfoot
\multirow{12}{*}{BART-base}
  & FT & -- & 0.02 & \textbf{33.47} & 10.49 & 44.16 & 35.33 \\
  & PB & 0  & 0.86$^\dagger$ & 24.68 & 11.20 & \textbf{47.63} & \textbf{49.44} \\
  & PB & 1  & 0.79$^\dagger$ & 28.37 & 10.90 & 47.06 & 47.42 \\
  & PB & 2  & 0.00 & 29.13 & 4.61 & -18.23 & 4.05 \\
  & PB & 3  & 0.66$^\dagger$ & 32.04 & 10.48 & 46.03 & 44.46 \\
  & PB & 4  & 0.63$^\dagger$ & 32.76 & 10.32 & 45.48 & 42.78 \\
  & PB & 5  & 0.63$^\dagger$ & 32.78 & 10.32 & 45.47 & 42.73 \\
  & PB & 6  & 0.76$^\dagger$ & 29.75 & 10.74 & 46.83 & 46.41 \\
  & PB & 7  & 0.65$^\dagger$ & 32.38 & 10.40 & 45.76 & 43.17 \\
  & PB & 8  & 0.66$^\dagger$ & 32.01 & 10.49 & 45.99 & 44.34 \\
  & PB & 9  & 0.59$^\dagger$ & 33.44 & 10.14 & 44.96 & 41.14 \\
  & PB & 10 & 0.73$^\dagger$ & 30.61 & 10.72 & 46.62 & 45.36 \\

\midrule

\multirow{12}{*}{BART-large}
  & FT & -- & 0.02 & \textbf{36.25} & 10.21 & 44.49 & 36.08 \\
  & PB & 0  & 0.84$^\dagger$ & 24.76 & 11.10 & \textbf{47.96} & \textbf{49.46} \\
  & PB & 1  & 0.77$^\dagger$ & 28.58 & 10.79 & 47.54 & 47.75 \\
  & PB & 2  & 0.00 & 30.92 & 4.38 & -17.65 & 3.88 \\
  & PB & 3  & 0.62$^\dagger$ & 32.40 & 10.35 & 46.37 & 44.89 \\
  & PB & 4  & 0.60$^\dagger$ & 32.96 & 10.28 & 45.99 & 43.45 \\
  & PB & 5  & 0.60$^\dagger$ & 33.00 & 10.28 & 45.96 & 43.40 \\
  & PB & 6  & 0.74$^\dagger$ & 30.21 & 10.66 & 47.27 & 46.48 \\
  & PB & 7  & 0.62$^\dagger$ & 32.50 & 10.36 & 46.12 & 43.91 \\
  & PB & 8  & 0.64$^\dagger$ & 32.21 & 10.42 & 46.28 & 44.77 \\
  & PB & 9  & 0.56$^\dagger$ & 33.77 & 10.08 & 45.23 & 41.96 \\
  & PB & 10 & 0.71$^\dagger$ & 31.10 & 10.63 & 47.08 & 45.52 \\

\midrule

\multirow{12}{*}{T5-base}
  & FT     & --  & 0.22           & 33.43 & 10.59 & \textbf{44.97} & \textbf{45.69} \\
  & PB     & 0   & 0.23           & 32.45 & 10.30 & 42.04 & 40.17 \\
  & PB     & 1   & 0.07           & 31.19 & 3.58 & -7.89 & 8.86 \\
  & PB     & 2   & 0.00           & 26.96 & 2.16 & -21.00 & 1.39 \\
  & PB     & 3   & 0.00           & 26.30 & 0.00 & -21.53 & 0.35 \\
  & PB     & 4   & 0.08           & 31.23 & 6.26 & -1.14 & 16.09 \\
  & PB     & 5   & 0.09           & 31.44 & 6.19 & -7.22 & 10.71 \\
  & PB     & 6   & 0.00           & 26.41 & 0.00 & -20.40 & 0.25 \\
  & PB     & 7   & 0.05           & 30.93 & 2.59 & -12.78 & 5.65 \\
  & PB     & 8   & 0.00           & 26.48 & 0.00 & -21.82 & 0.26 \\
  & PB     & 9   & 0.01           & \textbf{38.00} & 7.76 & 6.48 & 32.96 \\
  & PB     & 10  & 0.05           & 31.30 & 4.58 & -6.61 & 9.97 \\

\midrule

\multirow{12}{*}{T5-large}
  & FT     & --  & 0.20           & 33.22 & 10.53 & \textbf{44.87} & \textbf{48.53} \\
  & PB     & 0   & 0.01           & 35.97 & 1.02 & 19.60 & 3.08 \\
  & PB     & 1   & 0.00           & 34.05 & 2.27 & 21.15 & 5.13 \\
  & PB     & 2   & 0.00           & 36.61 & 5.49 & 20.03 & 3.88 \\
  & PB     & 3   & 0.00           & 34.55 & 2.22 & 20.05 & 7.55 \\
  & PB     & 4   & 0.00           & 37.54 & 3.59 & 17.76 & 4.59 \\
  & PB     & 5   & 0.00           & 35.49 & 5.18 & 16.13 & 2.71 \\
  & PB     & 6   & 0.00           & 34.74 & 2.17 & 17.74 & 4.13 \\
  & PB     & 7   & 0.00           & \textbf{38.22} & 3.86 & 10.99 & 3.54 \\
  & PB     & 8   & 0.00           & 37.59 & 0.36 & 16.81 & 2.45 \\
  & PB     & 9   & 0.00           & 34.69 & 3.91 & 18.92 & 6.56 \\
  & PB     & 10  & 0.00           & 34.45 & 3.54 & 19.08 & 3.55 \\

\midrule

\multirow{12}{*}{Flan-T5-base}
  & FT & -- & 0.04 & \textbf{36.24} & 9.68 & 42.63 & 38.44 \\
  & PB & 0  & 0.19 & 35.29 & 7.95 & 27.58 & 35.86 \\
  & PB & 1  & 0.69$^\dagger$ & 25.23 & 10.60 & 46.53 & 55.54 \\
  & PB & 2  & 0.34 & 31.53 & 9.98 & 43.32 & 57.70 \\
  & PB & 3  & 0.36 & 31.74 & 10.03 & 44.29 & 57.17 \\
  & PB & 4  & 0.36 & 33.35 & 9.54 & 44.52 & 59.48 \\
  & PB & 5  & 0.48 & 29.19 & 10.23 & 45.37 & 56.86 \\
  & PB & 6  & 0.27 & 33.88 & 9.73 & 44.17 & \textbf{60.03} \\
  & PB & 7  & 0.69$^\dagger$ & 25.22 & 10.56 & 45.98 & 55.31 \\
  & PB & 8  & 0.40 & 31.92 & 9.81 & 45.49 & 58.37 \\
  & PB & 9  & 0.41 & 31.55 & 9.83 & 45.50 & 57.20 \\
  & PB & 10 & 0.64$^\dagger$ & 26.22 & 10.56 & \textbf{46.57} & 54.33 \\

\midrule

\multirow{12}{*}{Flan-T5-large}
  & FT & -- & 0.06 & \textbf{36.62} & 9.38 & 43.29 & 38.79 \\
  & PB & 0  & 0.32 & 34.06 & 9.48 & 43.85 & 58.58 \\
  & PB & 1  & 0.46 & 30.90 & 10.24 & 46.74 & 56.00 \\
  & PB & 2  & 0.35 & 32.72 & 9.88 & 46.25 & 59.02 \\
  & PB & 3  & 0.34 & 33.12 & 9.81 & 45.78 & 58.21 \\
  & PB & 4  & 0.26 & 35.62 & 9.56 & 44.70 & \textbf{61.84} \\
  & PB & 5  & 0.46 & 31.89 & 10.10 & 46.90 & 58.20 \\
  & PB & 6  & 0.48 & 31.54 & 10.17 & \textbf{47.32} & 58.05 \\
  & PB & 7  & 0.74$^\dagger$ & 27.01 & 10.38 & 46.42 & 54.90 \\
  & PB & 8  & 0.69$^\dagger$ & 28.59 & 10.26 & 47.56 & 55.84 \\
  & PB & 9  & 0.49 & 31.05 & 10.04 & 47.15 & 58.90 \\
  & PB & 10 & 0.46 & 30.77 & 10.20 & 46.92 & 56.00 \\

\midrule

\multirow{12}{*}{Pegasus-large}
  & FT & -- & 0.45 & \textbf{28.56} & 11.09 & 49.88 & \textbf{54.39} \\
  & PB & 0  & 0.91$^\dagger$ & 22.23 & 11.85 & 52.38 & 54.54 \\
  & PB & 1  & 0.89$^\dagger$ & 22.64 & 11.95 & 52.06 & 54.06 \\
  & PB & 2  & 0.00 & 24.80 & 10.97 & 14.89 & 29.11 \\
  & PB & 3  & 0.00 & 22.08 & 16.20 & 37.63 & 39.41 \\
  & PB & 4  & 0.00 & 22.15 & 14.76 & \textbf{48.69} & 25.16 \\
  & PB & 5  & 0.00 & 21.73 & 14.30 & 48.26 & 32.20 \\
  & PB & 6  & 0.00 & 21.89 & 13.26 & 47.20 & 22.40 \\
  & PB & 7  & 0.84$^\dagger$ & 23.87 & 11.58 & 50.97 & 52.13 \\
  & PB & 8  & 0.00 & 24.92 & 16.69 & 33.08 & 12.35 \\
  & PB & 9  & 0.00 & 21.59 & 13.49 & 46.55 & 29.84 \\
  & PB & 10 & 0.87$^\dagger$ & 22.37 & 11.74 & 51.66 & 55.26 \\

\midrule

\noalign{\penalty -10000}
\multirow{12}{*}{Pegasus-xsum}
  & FT & --  & 0.45 & 27.09 & 11.39 & \textbf{49.94} & 55.35 \\
  & PB & 0  & 0.08 & 28.00 & 10.60 & 14.29 & 41.55 \\
  & PB & 1  & 0.09 & 28.04 & 10.19 & 14.55 & 45.02 \\
  & PB & 2  & 0.00 & 21.35 & 4.08 & 1.65 & 45.48 \\
  & PB & 3  & 0.02 & 24.09 & 6.81 & 8.80 & \textbf{60.22} \\
  & PB & 4  & 0.08 & 25.88 & 8.07 & 11.92 & 47.56 \\
  & PB & 5  & 0.09 & \textbf{28.15} & 8.75 & 18.70 & 44.24 \\
  & PB & 6  & 0.00 & 22.36 & 5.16 & 7.22 & 50.57 \\
  & PB & 7  & 0.05 & 26.03 & 8.16 & 12.38 & 52.58 \\
  & PB & 8  & 0.10 & 29.17 & 10.15 & 22.65 & 47.99 \\
  & PB & 9  & 0.08 & 25.47 & 8.42 & 12.20 & 49.12 \\
  & PB & 10 & 0.03 & 24.45 & 8.30 & 9.35 & 46.31 \\

\midrule

\multirow{12}{*}{\shortstack[l]{ProphetNet-large-\\uncased-cnndm}}
  & FT & -- & 0.11 (ci) & \textbf{36.45} & 9.33 & 40.18 & \textbf{55.99} \\
  & PB & 0  & 0.16 (ci) & 36.12 & 7.34 & 35.94 & 48.68 \\
  & PB & 1  & 0.22 (ci) & 33.69 & 9.66 & 38.24 & 48.74 \\
  & PB & 2  & 0.00 (ci) & 31.14 & 9.41 & 10.28 & 42.59 \\
  & PB & 3  & 0.06 (ci) & 35.02 & 6.54 & 17.62 & 45.64 \\
  & PB & 4  & 0.04 (ci) & 35.12 & 9.17 & 16.62 & 30.54 \\
  & PB & 5  & 0.03 (ci) & 35.20 & 10.12 & 22.50 & 29.62 \\
  & PB & 6  & 0.08 (ci) & 35.16 & 9.69 & 24.35 & 36.02 \\
  & PB & 7  & 0.13 (ci) & 33.59 & 8.56 & 34.64 & 39.71 \\
  & PB & 8  & 0.13 (ci) & 34.73 & 7.84 & 31.03 & 38.48 \\
  & PB & 9  & 0.05 (ci) & 34.24 & 8.52 & 23.74 & 48.74 \\
  & PB & 10 & 0.12 (ci) & 34.89 & 10.26 & \textbf{40.18} & 43.27 \\

\end{longtable}
\normalsize

\begin{longtable}{l l c c c c c c}
\caption{OneStopEnglish - Full per-prompt results}
\label{tab:ose-all-models-full}\\
\toprule
Model & Variant & Prompt\# & Identical ratio & SARI $\uparrow$ & FKGL $\downarrow$ & BERTScore $\uparrow$ & LENS $\uparrow$ \\
\midrule
\endfirsthead
\toprule
Model & Variant & Prompt\# & Identical ratio & SARI $\uparrow$ & FKGL $\downarrow$ & BERTScore $\uparrow$ & LENS $\uparrow$ \\
\midrule
\endhead
\midrule
\multicolumn{7}{r}{\emph{(table continues on next page)}} \\
\midrule
\endfoot
\bottomrule
\endlastfoot
\multirow{12}{*}{BART-base}
  & FT & --  & 0.00 & \textbf{37.45} & 8.08 & 75.46 & 41.18 \\
  & PB & 0  & 0.93$^{\dagger}$ & 27.36 & 9.26 & \textbf{81.10} & \textbf{59.80} \\
  & PB & 1  & 0.87$^{\dagger}$ & 28.59 & 9.07 & 80.27 & 57.83 \\
  & PB & 2  & 0.00 & 18.71 & 5.68 & -10.91 & 4.04 \\
  & PB & 3  & 0.72$^{\dagger}$ & 31.17 & 8.58 & 77.98 & 52.93 \\
  & PB & 4  & 0.67$^{\dagger}$ & 31.79 & 8.41 & 77.18 & 51.43 \\
  & PB & 5  & 0.67$^{\dagger}$ & 31.81 & 8.40 & 77.26 & 51.54 \\
  & PB & 6  & 0.82$^{\dagger}$ & 29.64 & 8.90 & 79.73 & 56.32 \\
  & PB & 7  & 0.70$^{\dagger}$ & 31.45 & 8.48 & 77.53 & 52.31 \\
  & PB & 8  & 0.72$^{\dagger}$ & 31.20 & 8.55 & 77.93 & 52.94 \\
  & PB & 9  & 0.62$^{\dagger}$ & 32.21 & 8.24 & 76.02 & 49.72 \\
  & PB & 10  & 0.80$^{\dagger}$ & 30.13 & 8.82 & 79.29 & 55.66 \\

\midrule

\multirow{12}{*}{BART-large}
  & FT & --  & 0.00 & \textbf{39.99} & 7.62 & 76.09 & 43.28 \\
  & PB & 0  & 0.90$^{\dagger}$ & 27.60 & 9.24 & \textbf{80.55} & \textbf{60.11} \\
  & PB & 1  & 0.86$^{\dagger}$ & 29.05 & 8.98 & 79.93 & 57.73 \\
  & PB & 2  & 0.00 & 33.50 & 4.56 & 28.85 & 11.45 \\
  & PB & 3  & 0.69$^{\dagger}$ & 31.28 & 8.37 & 76.76 & 52.39 \\
  & PB & 4  & 0.65$^{\dagger}$ & 31.95 & 8.35 & 76.42 & 51.09 \\
  & PB & 5  & 0.67$^{\dagger}$ & 31.82 & 8.38 & 77.10 & 51.56 \\
  & PB & 6  & 0.81$^{\dagger}$ & 29.98 & 8.88 & 79.19 & 56.38 \\
  & PB & 7  & 0.68$^{\dagger}$ & 31.55 & 8.47 & 77.13 & 52.40 \\
  & PB & 8  & 0.70$^{\dagger}$ & 31.29 & 8.49 & 77.58 & 52.76 \\
  & PB & 9  & 0.61$^{\dagger}$ & 32.32 & 8.24 & 75.50 & 49.69 \\
  & PB & 10  & 0.80$^{\dagger}$ & 30.38 & 8.79 & 79.28 & 55.66 \\

\midrule
\multirow{12}{*}{T5-base}
  & FT & --  & 0.31 & \textbf{37.51} & 8.14 & \textbf{76.70} & \textbf{56.97} \\
  & PB & 0  & 0.32 & 33.63 & 7.58 & 73.81 & 50.35 \\
  & PB & 1  & 0.08 & 26.50 & 1.97 & 8.14 & 13.77 \\
  & PB & 2  & 0.00 & 17.76 & 1.35 & -13.61 & 2.23 \\
  & PB & 3  & 0.00 & 16.36 & 0.00 & -15.38 & 0.22 \\
  & PB & 4  & 0.11 & 30.35 & 5.70 & 21.68 & 23.47 \\
  & PB & 5  & 0.09 & 26.90 & 4.63 & 8.33 & 14.12 \\
  & PB & 6  & 0.00 & 17.14 & 0.00 & -13.79 & 0.62 \\
  & PB & 7  & 0.05 & 25.87 & 2.17 & 5.53 & 12.28 \\
  & PB & 8  & 0.00 & 16.62 & 0.00 & -15.45 & 0.29 \\
  & PB & 9  & 0.02 & 30.87 & 6.94 & 17.85 & 30.72 \\
  & PB & 10  & 0.08 & 26.92 & 3.05 & 14.39 & 18.51 \\

\midrule

\multirow{12}{*}{T5-large}
  & FT & --  & 0.32 & \textbf{39.43} & 8.16 & \textbf{78.24} & \textbf{60.31} \\
  & PB & 0  & 0.02 & 35.62 & 0.00 & 42.61 & 5.09 \\
  & PB & 1  & 0.00 & 34.72 & 0.69 & 42.11 & 6.63 \\
  & PB & 2  & 0.00 & 35.76 & 4.13 & 37.07 & 4.92 \\
  & PB & 3  & 0.00 & 34.43 & 0.30 & 40.81 & 10.28 \\
  & PB & 4  & 0.00 & 35.10 & 1.67 & 36.17 & 6.02 \\
  & PB & 5  & 0.00 & 35.23 & 3.85 & 32.30 & 3.20 \\
  & PB & 6  & 0.00 & 34.70 & 0.10 & 36.15 & 5.35 \\
  & PB & 7  & 0.00 & 35.92 & 2.51 & 28.62 & 4.27 \\
  & PB & 8  & 0.00 & 35.12 & 0.00 & 35.92 & 3.17 \\
  & PB & 9  & 0.00 & 34.93 & 2.06 & 37.88 & 7.74 \\
  & PB & 10  & 0.00 & 35.39 & 2.01 & 38.77 & 4.65 \\

\midrule

\multirow{12}{*}{Flan-T5-base}
  & FT & --  & 0.02 & \textbf{37.73} & 7.57 & 72.84 & 45.97 \\
  & PB & 0  & 0.12 & 34.65 & 6.27 & 51.86 & 54.77 \\
  & PB & 1  & 0.69$^{\dagger}$ & 29.79 & 9.23 & \textbf{80.00} & 60.94 \\
  & PB & 2  & 0.37 & 33.37 & 8.67 & 76.31 & 61.02 \\
  & PB & 3  & 0.34 & 34.18 & 8.64 & 76.35 & 61.44 \\
  & PB & 4  & 0.21 & 37.24 & 7.79 & 74.68 & \textbf{63.78} \\
  & PB & 5  & 0.43 & 33.15 & 8.76 & 77.90 & 61.59 \\
  & PB & 6  & 0.22 & 36.65 & 7.83 & 74.94 & 63.16 \\
  & PB & 7  & 0.72$^{\dagger}$ & 29.56 & 9.14 & 79.60 & 60.05 \\
  & PB & 8  & 0.27 & 36.08 & 8.14 & 75.50 & 63.13 \\
  & PB & 9  & 0.30 & 35.52 & 8.26 & 76.25 & 62.64 \\
  & PB & 10  & 0.66$^{\dagger}$ & 30.11 & 9.13 & 79.25 & 60.52 \\

\midrule

\multirow{12}{*}{Flan-T5-large}
  & FT & --  & 0.04 & \textbf{38.38} & 7.64 & 74.23 & 46.45 \\
  & PB & 0  & 0.24 & 36.39 & 7.54 & 70.71 & 61.10 \\
  & PB & 1  & 0.35 & 33.62 & 8.79 & 77.58 & 61.51 \\
  & PB & 2  & 0.26 & 36.25 & 8.16 & 75.67 & 62.93 \\
  & PB & 3  & 0.25 & 37.12 & 8.08 & 76.69 & 62.68 \\
  & PB & 4  & 0.20 & 38.10 & 7.55 & 73.40 & \textbf{65.02} \\
  & PB & 5  & 0.39 & 34.05 & 8.69 & 77.64 & 62.45 \\
  & PB & 6  & 0.37 & 34.70 & 8.71 & 77.97 & 62.39 \\
  & PB & 7  & 0.72$^{\dagger}$ & 29.22 & 9.15 & \textbf{79.88} & 59.77 \\
  & PB & 8  & 0.51$^{\dagger}$ & 32.14 & 9.01 & 78.77 & 62.03 \\
  & PB & 9  & 0.39 & 33.94 & 8.70 & 77.69 & 62.86 \\
  & PB & 10  & 0.38 & 33.06 & 8.90 & 77.24 & 61.50 \\

\midrule

\noalign{\penalty -10000}
\multirow{12}{*}{Pegasus-large}
  & FT & --  & 0.41 & \textbf{36.89} & 8.42 & 77.79 & \textbf{61.13} \\
  & PB & 0  & 0.93$^{\dagger}$ & 26.95 & 9.49 & \textbf{80.89} & 59.32 \\
  & PB & 1  & 0.90$^{\dagger}$ & 27.28 & 9.40 & 80.52 & 58.41 \\
  & PB & 2  & 0.00 & 18.76 & 9.67 & 15.46 & 28.97 \\
  & PB & 3  & 0.00 & 27.45 & 14.53 & 55.60 & 38.36 \\
  & PB & 4  & 0.00 & 26.98 & 12.72 & 64.42 & 24.61 \\
  & PB & 5  & 0.00 & 28.03 & 12.27 & 63.78 & 33.31 \\
  & PB & 6  & 0.00 & 26.99 & 10.75 & 63.65 & 23.98 \\
  & PB & 7  & 0.78$^{\dagger}$ & 30.04 & 8.97 & 78.95 & 54.54 \\
  & PB & 8  & 0.00 & 29.43 & 16.39 & 38.96 & 12.42 \\
  & PB & 9  & 0.00 & 27.13 & 11.04 & 63.17 & 30.85 \\
  & PB & 10  & 0.87$^{\dagger}$ & 28.22 & 9.27 & 80.04 & 57.35 \\

\midrule

\multirow{12}{*}{Pegasus-xsum}
  & FT & --  & 0.40 & \textbf{37.07} & 8.66 & \textbf{77.77} & 60.97 \\
  & PB & 0  & 0.02 & 23.08 & 8.94 & 17.98 & 56.81 \\
  & PB & 1  & 0.09 & 27.63 & 8.12 & 28.99 & 59.72 \\
  & PB & 2  & 0.00 & 17.70 & 3.42 & -1.11 & 55.63 \\
  & PB & 3  & 0.02 & 21.39 & 5.51 & 9.74 & \textbf{65.54} \\
  & PB & 4  & 0.01 & 21.59 & 6.27 & 5.39 & 57.36 \\
  & PB & 5  & 0.06 & 25.66 & 6.47 & 20.65 & 58.89 \\
  & PB & 6  & 0.00 & 19.25 & 4.10 & 5.41 & 61.33 \\
  & PB & 7  & 0.00 & 19.74 & 6.61 & 8.71 & 61.78 \\
  & PB & 8  & 0.06 & 28.31 & 8.31 & 27.34 & 57.04 \\
  & PB & 9  & 0.04 & 22.72 & 6.27 & 12.09 & 59.99 \\
  & PB & 10  & 0.02 & 22.08 & 7.41 & 9.91 & 58.54 \\

\midrule

\multirow{12}{*}{\shortstack[l]{ProphetNet-large-\\uncased-cnndm}}
  & FT & --  & 0.13 (ci) & \textbf{39.17} & 7.00 & \textbf{65.22} & \textbf{61.53} \\
  & PB & 0  & 0.07 (ci) & 35.23 & 3.73 & 51.04 & 46.06 \\
  & PB & 1  & 0.14 (ci) & 35.60 & 7.05 & 58.27 & 50.91 \\
  & PB & 2  & 0.00 (ci) & 22.73 & 8.19 & 9.35 & 44.04 \\
  & PB & 3  & 0.02 (ci) & 29.95 & 4.85 & 21.86 & 47.15 \\
  & PB & 4  & 0.02 (ci) & 30.65 & 7.20 & 14.46 & 30.03 \\
  & PB & 5  & 0.01 (ci) & 33.53 & 7.77 & 28.93 & 30.17 \\
  & PB & 6  & 0.03 (ci) & 33.90 & 7.38 & 29.34 & 36.40 \\
  & PB & 7  & 0.05 (ci) & 35.23 & 5.95 & 47.03 & 39.65 \\
  & PB & 8  & 0.07 (ci) & 35.04 & 5.45 & 38.11 & 36.04 \\
  & PB & 9  & 0.02 (ci) & 32.65 & 6.24 & 25.75 & 48.16 \\
  & PB & 10  & 0.04 (ci) & 34.85 & 7.78 & 43.84 & 42.71 \\

\end{longtable}

\newpage
\section{Human Evaluation Details}
\label{app:human_eval}

\paragraph{Instructions.} 
Participants were instructed as follows:
\begin{quote}
In the following questionnaire you are requested to be the judge of a text simplification task. 
Each section will contain a \textit{source} sentence followed by two attempts to simplify it. 
One or both of the sentences may be a poor simplification. 
Please select the better one according to your judgment. 
If both appear the same to you, select ``same'' (but try to avoid this option as much as possible).
\end{quote}

\begin{figure}[h]
    \centering
    \includegraphics[width=0.9\linewidth]{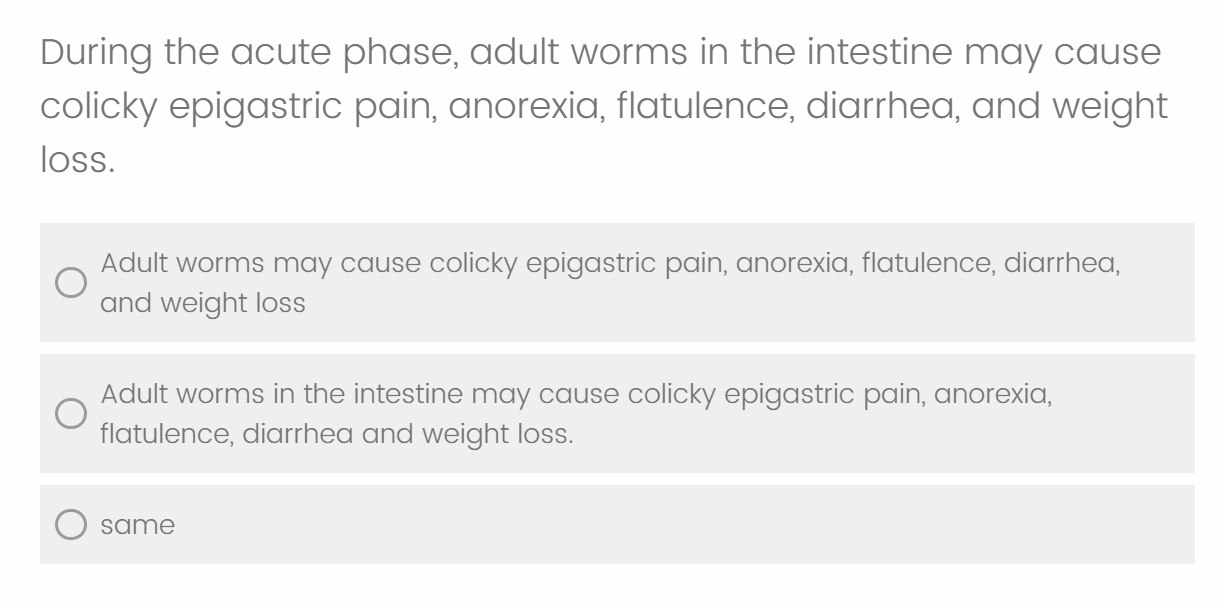}
    \caption{Screenshot of the human evaluation interface in Qualtrics. 
    Raters compared two candidate simplifications (order randomized) and could also select ``same''.}
    \label{fig:qualtrics-interface}
\end{figure}

\paragraph{Interface.} 
Figure~\ref{fig:qualtrics-interface} shows an example of the Qualtrics interface as presented to raters. 
The source sentence was displayed at the top, with the two candidate simplifications below in randomized order, followed by the ``same'' option.

\paragraph{Qualtrics Survey Link}
\label{sec:qualtrics}

The human evaluation questionnaire is available at:  
\url{https://qualtricsxmrlzdwvxhq.qualtrics.com/jfe/form/SV_enYDSTxInoXLEVw}

\end{document}